\documentclass[journal]{IEEEtran}

\ifCLASSOPTIONcompsoc
  \usepackage[nocompress]{cite}
\else
  \usepackage{cite}
\fi

\usepackage[usenames,dvipsnames]{color}

\usepackage{pgfplots}
\pgfplotsset{compat=newest}

\usepackage{enumitem}
\usepackage{epsfig}
\usepackage{graphicx}
\usepackage{amsmath}
\usepackage{amsfonts}
\usepackage{amssymb}
\usepackage{array}
\usepackage{multirow}
\usepackage{rotating}
\usepackage{epstopdf}
\usepackage{caption}
\usepackage{subcaption}
\usepackage{xcolor}

\usepackage[makeroom]{cancel}

\usepackage{url}
\usepackage{afterpage}
\usepackage{booktabs}

\newcommand{\rulesep}{\unskip\ \vrule\ }

\begin{document}


\title{General Framework to Evaluate Unlinkability\\in Biometric Template Protection Systems}
%

\author{Marta~Gomez-Barrero, Javier~Galbally, Christian~Rathgeb, Christoph~Busch
\thanks{M. Gomez-Barrero, C. Rathgeb and C. Busch are with the da/sec - Biometrics and Internet Security Research Group, Hochschule Darmstadt, Germany (e-mail: \{marta.gomez-barrero,christian.rathgeb,christoph.busch\}@h-da.de).}
\thanks{J. Galbally is with the European Commission - DG-Joint Research Centre, E.3, Italy (e-mail: javier.galbally@ec.europa.eu).}}%

%
%

\markboth{}
{}
%



\maketitle

\begin{abstract}
The wide deployment of biometric recognition systems in the last two decades has raised privacy concerns regarding the storage and use of biometric data. As a consequence, the ISO/IEC 24745 international standard on biometric information protection has established two main requirements for protecting biometric templates: irreversibility and unlinkability. Numerous efforts have been directed to the development and analysis of irreversible templates. However, there is still no systematic quantitative manner to analyse the unlinkability of such templates. In this paper we address this shortcoming by proposing a new general framework for the evaluation of biometric templates' unlinkability. To illustrate the potential of the approach, it is applied to assess the unlinkability of four state-of-the-art techniques for biometric template protection: biometric salting, Bloom filters, Homomorphic Encryption and block re-mapping. For the last technique, the proposed framework is compared with other existing metrics to show its advantages.
\end{abstract}


%
\IEEEpeerreviewmaketitle

\setlength{\tabcolsep}{4.0pt}

\section{Introduction}
\label{sec:intro}


Biometrics refers to automated recognition of individuals based on their biological or behavioural characteristics, such as iris or signature \cite{jain07nature}. Its advantages over traditional authentication methods (e.g., no need to carry tokens or remember passwords, harder to circumvent or stronger link between the subject and the action or event), have allowed a wide deployment of biometric systems in the last decade, including large-scale national and international initiatives such as the Unique ID program of the Indian government \cite{indianUID} or the Smart Borders project of the European Comission \cite{SmartBorders}. However, unprotected storage of biometric reference templates poses severe privacy threats, e.g. identity theft, cross-matching or limited renewability. In fact, biometric data are defined as sensitive data within the European Union (EU) General Data Protection Regulation 2016/679 \cite{euregulation16}, which means that the use of these data is subjected to the right of privacy preservation. 

Considering those privacy issues, biometric template protection schemes have been developed in the last two decades \cite{campisi13secPrivacyBio,patel15CancelableBioSurvey}, and several standardization efforts \cite{ISO-IEC-24745:2011,ISO-IEC-30136-2017,rane14standardBPT} have been directed to this topic. Biometric template protection schemes are commonly categorized as \emph{biometric cryptosystems}, also referred to as helper data schemes, and \emph{cancelable biometrics}, also referred to as feature transformation approaches. Biometric cryptosystems are designed to securely bind a digital key to a biometric characteristic or generate a digital key from a biometric signal \cite{BUludag04a}. Cancelable biometrics consist of intentional, repeatable distortions of biometric signals based on transforms that provide a comparison of biometric templates in the transformed domain \cite{BRatha01a,patel15CancelableBioSurvey}.

As defined in the ISO/IEC International Standard 24745 on biometric information protection \cite{ISO-IEC-24745:2011}, in order to protect the privacy of the individuals, \lq\lq knowledge of the transformed biometric reference cannot be used to determine any information about the generating biometric sample(s) or features'', which makes clear reference to the necessity of storing \emph{irreversible} biometric templates. But not only that, the ISO/IEC standard continues by stating \lq\lq[... and] the stored biometric references should not be linkable across applications or databases''. That is, protected templates are not only required to be \textit{irreversible}, but also \textit{unlinkable}. Only by fulfilling both requirements can we grant the privacy to which subjects are entitled.




As stated in \cite{rane14standardBPT}, a standardised benchmark protocol for biometric template protection schemes, in terms of recognition accuracy, security and privacy, is necessary for a further deployment of biometric verification systems. However, whereas the irreversibility of protected templates or the accuracy degradation with respect to unprotected systems have been thoroughly analysed in the literature, little attention has been paid to the objective evaluation of their unlinkability \cite{Rathgeb11e}. In fact, there is still no general metric, protocol or framework to assess, in an objective way, the unlinkability of biometric templates in order to be able to establish a fair benchmark for the performance of different protection algorithms. Due to the limitations of the current proposals to evaluate this property (for more details, see Sect.~\ref{sec:related}), no standardised metric has been included in the current ISO/IEC 30136 project on performance testing of biometric template protection schemes \cite{ISO-IEC-30136-2017}.

To tackle the aforementioned issue, in the present article we propose a general framework that addresses this need. The novel methodology is partially inspired by the initial ideas presented by Ferrara \textit{et al.} in \cite{Ferrara14a}, where the authors consider three sets of scores distributions resulting from the comparison of templates enrolled in different applications using different application-specific parameters. However, probably due to the lack of an appropriate measure to benchmark them, no objective numerical analysis was carried out in their work, only a visual analysis between the score distributions. In this context, we provide two different metrics that enable the quantitative assessment of templates' unlinkability and the objective benchmark among systems. The two metrics are:  $i)$ a local score-wise measure, $\mathrm{D}_{\leftrightarrow} \left( s \right)$, based on the likelihood ratio between the score distributions; and $ii)$ a global measure, $\mathrm{D}_{\leftrightarrow}^{\mathit{sys}} $, independent of the score domain, and thereby allowing a fairer benchmark of the overall systems' unlinkability. Both measures yield values in a closed range and build upon the solid theory behind likelihood ratios for calibration of scores or interpretation of evidence in forensic environments \cite{champod2000LRspeaker,bazen2004LRverification,joaquin06LRspeaker,ali2013LRface,Ramos2017LRfp}. Furthermore, they are defined for the complete domain of scores in order to allow a more straightforward benchmark of different schemes, useful for instance for evaluating systems in competitions or for commercial purposes.

The rest of the paper is structured as follows. Previous works on unlinkability analysis of particular biometric template protection systems are summarised in Sect.~\ref{sec:related}. General concepts on biometric template protection and unlinkability are included in Sect.~\ref{sec:btp}. The new framework for the systematic analysis of the unlinkability of biometric template protection systems is described in Sect.~\ref{sec:metric}. A protocol for the linkability analysis of biometric templates is provided in Sect.~\ref{sec:protocol}. Then, the framework is applied to evaluate the unlinkability of four different state-of-the-art template protection schemes in Sect.~\ref{sec:results}, and conclusions are drawn in Sect.~\ref{sec:conc}
\section{Related Works}
\label{sec:related}

As mentioned in Sect.~\ref{sec:intro}, few works related to biometric template protection (BTP) have included some type of analysis of templates' unlinkability. These are summarised in Table~\ref{tab:sota}. For instance, Linnartz \textit{et al.} analyse in \cite{linnartz2003shieldingFunct} the information leaked by the stored reference templates and the danger posed by replay attacks. On the other hand, Dodis \textit{et al.} show that public data can disclose information about the original biometric sample, and measure this leak in terms of the average min entropy \cite{dodis2004bioKeys}. In addition, Buhan \textit{et al.} generalise the concept of fuzzy extractors to continuous distributions in \cite{buhan2007fuzzyContinuous}, establishing as well a relationship between the False Match Rate (FMR) of the system and the min entropy of the templates. 

From a more general perspective, a framework for the security and privacy analysis of biometric systems is developed in \cite{simoens2012framework}. In that work, a general system, involving four logical entities (i.e., sensor, server, database and comparator), is defined, and the security of the system is analysed assuming different adversary models where some of the aforementioned entities may be malicious. While the information flow and the ability to reconstruct templates is analysed, no metric or protocol is provided to measure the unlinkability of a given system.

Other works have proposed specific cross-matching attacks (which can also be considered as a \textit{linkage function}, as will be seen in Sect.~\ref{sec:btp}) in order to link templates produced by several template protection schemes, and analysed the threat posed in terms of their success chances. For instance, in \cite{bringer15secAnalysisBF}, the distributions of dissimilarity scores for normal and attacking comparisons are depicted, but no quantitative measure of the linkability is provided. In \cite{kholmatov2008corrAttackFV}, a cross-matching attack against fuzzy vault schemes is proposed. However, in this case, due to the time consuming nature of the attack, only ten vaults are analysed, reporting a success rate of the attack of 40\% (which could be understood as a very attack-specific linkability metric). 

For the particular case of biometric cryptosystems, the unlinkability property has been frequently referred to as \textit{indistinguishability} \cite{simoens2009indistinguishabilityBioSketches,buhan2009indistinguishabilityBCS,buhan2010indistinguishabilityFuzzy}, being its definition identical to the unlinkability definition included in \cite{ISO-IEC-24745:2011,ISO-IEC-30136-2017} (see Sect.~\ref{sec:btp}). In their work \cite{simoens2009indistinguishabilityBioSketches}, Simoens \textit{et al.}, following a cryptographic perspective, analyse the unlinkability of biometric sketches playing the so-called \textit{indistinguishability game}. Assuming the attacker or adversary has obtained a set of sketches, his objective is to identify related sketches. In order to measure his success chances, they estimate his advantage over a random guess. This is done by setting bounds or computing probabilities which rely on the Error Correcting Codes (ECC) theory, which is fundamental to the development of fuzzy sketches. Therefore, such an analysis can not be extrapolated to systems based, for instance, on cancelable biometrics approaches. In addition, one of the main limitations of the theoretical derivation is the assumption of handling uniform data, which does not hold for biometric data. 

A similar approach is followed in \cite{buhan2009indistinguishabilityBCS} for the analysis of Quantization Index Modulation (QIM) biometric cryptosystems. Since \lq\lq the concept describes the advantage of an attacker with respect to a perfect indistinguishable system'', which is hard to achieve in practice due to the inherent correlation of biometric data, the authors propose an alternative practical evaluation in which they benchmark Equal Error Rates (EER) obtained for regular comparisons of templates protected with a single key (i.e., recognition accuracy analysis) against comparisons of templates protected with different keys (i.e., unlinkability analysis). For such comparisons, they use the same distinguisher function used for the indistinguishability game. The main drawback of this practical approach is that, even if an increase of the EER implies some degree of unlinkability of the system (i.e., the system loses discriminative power when different keys are used to protect the templates), such unlinkability increase is not quantified. 

\begin{table*}[t]
\begin{small}
\begin{center}

\caption{Summary of the most relevant methodologies for unlinkability assessment and their main properties: $i)$ whether they are general enough to be applied to different BTP schemes, $ii)$ whether they regard linkability as a binary or continuous characteristic, $iii)$ what kind of metric is proposed, $iv)$ whether a quantitative metric is proposed and $v)$ what assumptions are made. Drawbacks are highlighted in bold letters.}\label{tab:sota}
\centering
\begin{tabular}{llccccc}
\toprule
\textbf{Ref.} & \phantom{c} & \textbf{General?} & \textbf{Binary / Continuous}  & \textbf{Metric} & \textbf{Quantitative?} & \textbf{Assumptions}\\ \midrule
\cite{bringer15secAnalysisBF} & &Yes & - & Score distributions & \textbf{No} & - \\
\cite{kholmatov2008corrAttackFV} & &Yes & \textbf{Binary} & Success rate & Yes & -\\
\cite{simoens2009indistinguishabilityBioSketches} & & \textbf{No}: Biometric sketches &\textbf{Binary} & Adversary's advantage & Yes & \textbf{Uniform data}  \\
\cite{buhan2009indistinguishabilityBCS} & & \textbf{No}: QIM &\textbf{Binary} & Adversary's advantage & Yes & \textbf{Uniform data}  \\
\cite{buhan2009indistinguishabilityBCS} & & Yes & \textbf{Binary} & \textbf{Accuracy oriented curves (ROC)} & \textbf{No} & -  \\
\cite{buhan2010indistinguishabilityFuzzy} & & \textbf{No}: Fuzzy scheme & Continuous & Adversary's advantage & Yes &  \textbf{Uniform data}  \\
\cite{kelkboom2011crossmatchFuzzy} & & Yes & \textbf{Binary} & \textbf{Accuracy oriented curves (ROC)} & \textbf{No} & -  \\
\cite{nagar2010btpSecurity} & & Yes & \textbf{Binary} & \textbf{Accuracy oriented curves (DET)} & \textbf{No} & -  \\
\cite{piciucco2016cancelableFV} & & Yes & \textbf{Binary} & \textbf{Accuracy oriented curves (DET)} & \textbf{No} & -  \\
\cite{rua12BTPandUBMsign} & & Yes & Continuous & \textbf{Accuracy oriented curves (CMC)} & Yes & - \\
\cite{Ferrara14a} & &Yes & - & Score distributions & \textbf{No} & - \\
\cite{wang2014cancelableFpConv} & &Yes & - & Score distributions & \textbf{No} & - \\
\bottomrule
\end{tabular}\vspace{-0.5cm}
\end{center}
\end{small}
\end{table*}

In \cite{buhan2010indistinguishabilityFuzzy}, Buhan \textit{et al.} consider for the first time a \textit{continuous} value, instead of binary, for the linkability of two given templates for a fuzzy scheme. In other words, unlinkability is regarded as a continuous property, being possible to assign different \textit{degrees of linkability} to the templates. To that end, they introduce \lq\lq a \textit{classification function} to decide whether the query sketch and the target sketch are related''. This function is hence used to link templates. Then, the advantage of an eventual attacker over a random guess is theoretically modelled, and a relationship is established with the False Match Rate (FMR) and False Non-Match Rate (FNMR) of the system. It should be noted that similar unrealistic assumptions to those of \cite{simoens2009indistinguishabilityBioSketches} are made in the theoretical derivation. In addition, the FMR and FNMR are estimated in terms of the verification function of the system, which yields less discriminative information in terms of unlinkability than the classification function used by the attacker.

Kelkboom \textit{et al.} present a practical evaluation of the robustness of a fuzzy commitment scheme to several cross-matching attacks in \cite{kelkboom2011crossmatchFuzzy}. To that end, and following the approach presented in \cite{buhan2009indistinguishabilityBCS} for the EERs, the Receiving Operating Characteristic (ROC) curves of the system for the accuracy and unlinkability evaluations are compared: if the accuracy shown by the ROC curves decreases, then the system is assumed to be unlinkable. Again, a relationship of the vulnerability to the linkability attack and the FMR and FNMR is established, assuming the extracted bits to be independent with equal bit-error probability, a fact that does not always hold in biometric data due to its inherent correlation. In a similar manner, Nagar \textit{et al.} analyse the unlinkability of a biometric template protection scheme in \cite{nagar2010btpSecurity}, where the error rates for the unlinkability analysis are referred to as Cross Match Rate (CMR) and False Cross Match Rate (FCMR).


Following this same concept of using accuracy curves, Piciucco \textit{et al.} define in \cite{piciucco2016cancelableFV} the Renewable Template Matching Rate (RTMR) as the percentage of correctly linked templates. For the evaluation, a Detection Error Trade-off (DET)-like curve is depicted, where FNMR is shown for each RTMR for a subset of the database used in the experimental evaluation. In this case, mated scores for the FNMR are computed comparing templates protected with the \textit{same key} and extracted from the same instance, whereas non-mated scores for the RTMR are obtained from the comparison of templates protected with \textit{different keys} and extracted from different instances. The authors point out that the FNMR \textit{vs}. RTMR curve should be similar to the common DET curve (FNMR \textit{vs}. FMR) to indicate that matching templates protected with different keys is at least as hard as achieving a false match with templates obtained from different subjects. Therefore, if the curves are visually similar, unlinkability is obtained.


In \cite{rua12BTPandUBMsign}, a linkage function based on Principal Components Regression is defined for fuzzy commitment systems using Universal Background Models (UBM). Building upon this linkage function, the eventual attacker decides which is the identity hidden by both templates. In order to measure linkability, the authors focus on the probability that the attacker chooses the correct identity in a top-N list, and how this probability deviates from a random guess (full unlinkability). In other words, the authors analyse the variation of the Cumulative Match Curves (CMC) used to evaluate the accuracy of identification biometric systems. While this is a clever approach to solve the problem with some highly desirable properties (i.e., it is general, continuous and provides a quantitative measure), it still presents some limitations such as: $i)$ it does not give the linkability level for a given score (i.e., what is the probability that, given score $s_0$, the two templates stem from the same subject?); and $ii)$ it does not provide one unique general linkability measure for the whole system, but a different value for each size N of the identities list.

In contrast to this DET, CMC or ROC approach, which can hide how linkable templates are for specific subsets of scores (see Sect.~\ref{sec:exp:comp}), \cite{Ferrara14a,wang2014cancelableFpConv} have addressed the problem of unlinkability evaluation directly analysing the similarity scores. In \cite{Ferrara14a}, Ferrara \textit{et al.} consider three sets of scores distributions resulting from the comparison of templates enrolled in different applications using different secret keys. More specifically, templates were extracted from $i)$ the same sample of a given instance, $ii)$ different samples of the same instance or $iii)$ samples of different instances. The analysis in \cite{wang2014cancelableFpConv} focuses only on the last two distributions, which are the most likely to occur in a real-world scenario. In both works, only a visual benchmark between the score distributions is carried out, presumably due to a lack of an appropriate metric. Traditionally, such difference between probability densities has been estimated in terms of the Kullback-Leibler ($\mathit{KL}$) divergence \cite{kullback51KLdivergence} between two discrete distributions, $P$ and $Q$, which is defined as:
\begin{equation}
D_{\mathit{KL}} \left( P || Q \right) = \sum_s P(s) \ln \left( \frac{P(s)}{Q(s)} \right)
\end{equation}
where $D_{\mathit{KL}} \ge 0$, and $D_{\mathit{KL}} = 0$ holds iff $P \simeq Q$, i.e. the smaller $D_{\mathit{KL}}$, the higher the similarity between distributions. However, this measure presents several limitations for the unlinkability analysis due to three main reasons: $i)$ it gives only an overall measure of the unlinkability of the system, not being possible to measure the level of unlinkability for different domains of the linkage scores, $ii)$ it is not bounded, thus making it difficult to benchmark the unlinkability of different systems, and $iii)$ it is not defined for $Q(s) = 0$ if $P(s) \ne 0$, hence not taking into account important ranges of scores, or not being at all defined for fully separable distributions. 





In summary, although being all valuable contributions, the aforementioned articles share some common shortcomings, as highlighted in red in Table~\ref{tab:sota}: 
\begin{itemize}
\item Unrealistic assumptions on uniformity of biometric data \cite{simoens2009indistinguishabilityBioSketches,buhan2009indistinguishabilityBCS,buhan2010indistinguishabilityFuzzy}. 

\item Non general approaches: many of the methods proposed are developed for a specific system or template protection technique but cannot be applied to others. As a consequence, it is difficult or simply not possible to use them to benchmark different systems \cite{simoens2009indistinguishabilityBioSketches,buhan2009indistinguishabilityBCS,buhan2010indistinguishabilityFuzzy}.

\item Linkability is regarded as a binary decision, either templates are fully linkable, or fully unlinkable, and no \textit{degree} of linkability is considered \cite{kholmatov2008corrAttackFV,simoens2009indistinguishabilityBioSketches,buhan2009indistinguishabilityBCS,kelkboom2011crossmatchFuzzy,nagar2010btpSecurity,piciucco2016cancelableFV}.

\item Some approaches only suggest how linkability could be measured, but do not give any quantitative measure, just a subjective analysis \cite{Ferrara14a,bringer15secAnalysisBF,wang2014cancelableFpConv}.

\item Other methods recommend the use of metrics employed for verification accuracy evaluations, such as DET or ROC curves, not suitable for the linkability evaluation, as it will be shown in Sect.~\ref{sec:exp:comp} \cite{buhan2009indistinguishabilityBCS,buhan2010indistinguishabilityFuzzy,kelkboom2011crossmatchFuzzy,nagar2010btpSecurity,piciucco2016cancelableFV,rua12BTPandUBMsign}.


\end{itemize}


The proposed framework tackles such limitations and offers the following advantages:

\begin{itemize}
\item No assumptions are made on the data, neither on bits' independence (as understood in entropy studies)  nor on uniformity.


\item Only a classification function, named as \lq\lq linkage function'', is assumed to exist, in order to assess the non-binary nature of the unlinkability property \cite{buhan2010indistinguishabilityFuzzy}. In addition, to make the framework as general as possible, the function can take arguments in both the original (i.e., unprotected) or the protected domain, as in \cite{nagar2010btpSecurity}.

\item The proposed metrics evaluate linkability based on score distributions obtained from the linkage function, independently of what the linkage function is. This allows for a general metric, since it can be computed for any Lebesgue integrable linkage function.


\item Not only a global measure for the unlinkability of the templates is provided, but also a local measure for each linkage score, in order to allow a more thorough evaluation.

\item Using always the same metric for different linkage functions has the advantage of allowing to monitor the changes in a system's linkability when different functions are used to compare the templates.

\end{itemize}
\section{Concepts on Biometric Template Protection and Unlinkability}
\label{sec:btp}

Throughout the article we will use the Harmonized Biometric Vocabulary (HBV) defined in the ISO/IEC 2382-37 \cite{ISO-IEC-2382-37:2012}. For any clarification on the concepts used we refer the reader to the mentioned standard\footnote{Available at \url{http://standards.iso.org/ittf/PubliclyAvailableStandards/c055194_ISOIEC_2382-37_2012.zip} or \url{http://www.christoph-busch.de/standards.html}}. Given that they are often used throughout the article, for the sake of clarity, we will only include here the next definitions:
\begin{itemize}

\item \textit{Biometric characteristic}: \lq\lq biological and behavioural characteristic of an individual from which distinguishing, repeatable biometric features can be extracted for the purpose of biometric recognition''. For example, a fingerprint or an iris are two different biometric characteristics.

\item\textit{Biometric instance}: for some characteristics, an individual possesses several instances. For example, the right index fingerprint is a different instance from the left thumb, even if they serve to identify the same data subject. 

\item \textit{Mated samples}: \lq\lq paired biometric probe and biometric reference that are from the same biometric characteristic of the same biometric data subject''. For example, two fingerprint samples from the same right index finger.

\item \textit{Non-mated samples}: \lq\lq paired biometric probe and biometric reference that are not from the same biometric instance''. For example, two fingerprint samples from different fingers.

\end{itemize}

%
%
%

Within ISO/IEC IS 24745 \cite{ISO-IEC-24745:2011}, unlinkability is defined as \lq\lq a property of two or more biometric references that they cannot be linked to each other or to the subject(s) from which they were derived''. The challenge is hence to determine whether two protected templates, $\mathbf{T}_1$ and $\mathbf{T}_2$, enrolled in two different applications, conceal the same biometric instance (i.e., they represent different samples of biometric data extracted from the same biometric instance - e.g., the same left index finger). 

From an analytic perspective, and taking into account the works described in Sect.~\ref{sec:related}, the unlinkability definition given in the ISO/IEC IS 24745 presented above can be reformulated as a gradual property of the templates:

\vspace{0.2cm}
\noindent\fbox{%
    \parbox{0.97\linewidth}{%
       \textbf{Definition of linkability}: two templates are \textit{fully linkable} if there exists some method to decide that they were extracted, with all certainty, from the same biometric instance. Two templates are \textit{linkable to a certain degree} if there exists some method to decide that it is more likely that they were extracted from the same instance than from different instances.
    }%
}
\vspace{0.2cm}

Given the aforementioned definition of linkability, it follows that this property is fully related to the \textit{method} (i.e., linkage function) used to decide if two templates stem from the same instance. In fact, the different cross-matching attacks developed in the works described in Sect.~\ref{sec:related} are particular examples of linkage functions. 

%

\begin{figure}[t]
\centering
 \centerline{\includegraphics[width=.99\linewidth]{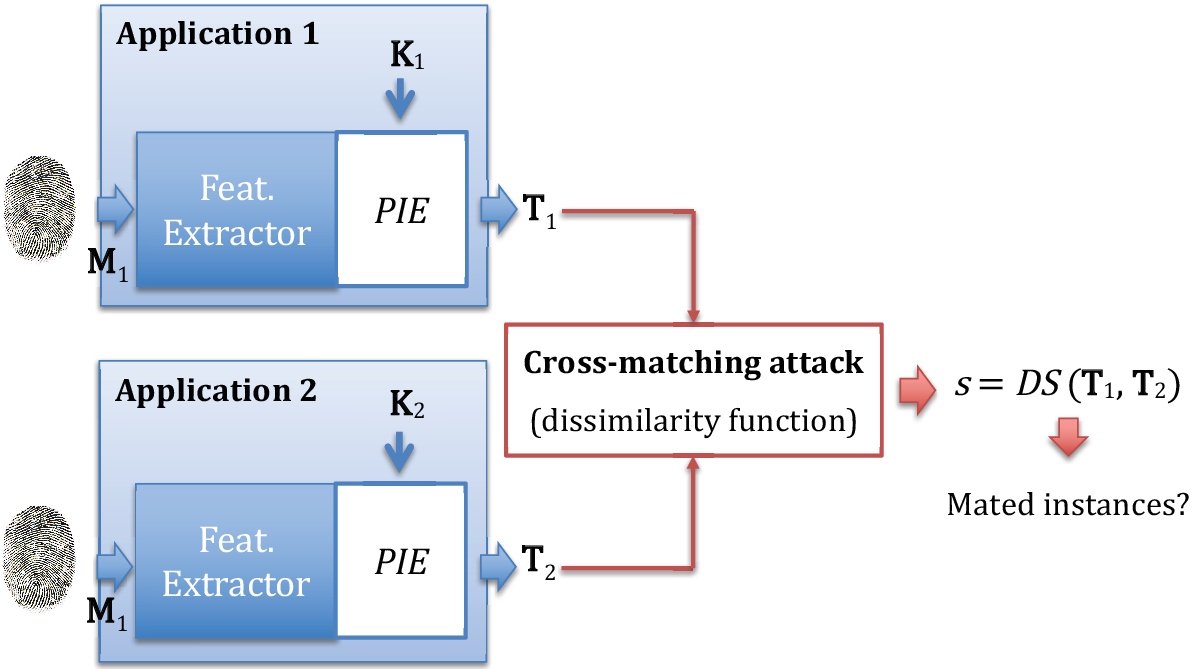}}
\caption{General diagram for linking templates. For two different applications, at enrollment the $\mathit{PIE}$ takes as input the reference biometric sample $\mathbf{M}$ and the corresponding secret key $\mathbf{K}_i$ to generate the protected templates $\mathbf{T}_i = \mathit{PIE}\left( \mathbf{M}, \mathbf{K}_i \right)$. Then, taking as input  $\mathbf{T}_1$ and $\mathbf{T}_2$, a linkage score between them, $s = LS\left( \mathbf{T}_1, \mathbf{T}_2\right)$, can be generated and used to decide whether both templates conceal mated instances. } \label{fig:attack}\vspace{-0.5cm}
\end{figure}

The scenario to be analysed is depicted in more detail in Fig.~\ref{fig:attack}. Given two applications, the biometric reference sample $\mathbf{M}$ is presented to the each of them during enrolment. Features are extracted and the \emph{Pseudonymous Identifier Encoder} ($\mathit{PIE}$), taking as input both the sample $\mathbf{M}$ and a secret key $\mathbf{K}_i$, with $i = \lbrace 1, 2 \rbrace$, computes the corresponding reference template, $\mathbf{T}_i = \mathit{PIE}\left( \mathbf{M}, \mathbf{K}_i \right)$. 

Following the aforementioned definition and the approach presented in \cite{buhan2010indistinguishabilityFuzzy}, where a classification function is used to evaluate unlinkability in a continuous manner, a \textit{linkage function} will provide a linkage score between the analysed templates: $s = LS \left( \mathbf{T}_1, \mathbf{T}_2 \right)$. Such function might be the similarity score computed by the BTP scheme or any other function exploiting a vulnerability of the system. This score will be used to determine whether both templates conceal the same instance. In other words, a BTP scheme is considered linkable \textit{to some degree} if, given a score $s = \mathit{LS}\left( \mathbf{T}_1, \mathbf{T}_2\right)$, the likelihood that both templates $\mathbf{T}_1$ and $\mathbf{T}_2$ conceal the same instance is larger than the likelihood that they conceal different instances.

In summary, the proposed framework to evaluate unlinkability assumes the following:
\begin{itemize}
\item The existence of a given linkage function $\mathit{LS}$ that, given two templates $\mathbf{T}_1 $ and $\mathbf{T}_2 $, produces a score $s = \mathit{LS}\left( \mathbf{T}_1, \mathbf{T}_2\right)$. This linkage function can be the $\mathit{PIC}$ (Pseudonymous Identifier Comparator) of the original system or some other function thought to exploit some vulnerability of the system. The only requisite is that it produces continuous normalizable scores.

\item Access to the linkage score $s$ between templates $\mathbf{T}_1 $ and $\mathbf{T}_2 $, $s = \mathit{LS}\left( \mathbf{T}_1, \mathbf{T}_2\right)$.
\end{itemize}

\begin{figure*}[t]
\centering
\begin{subfigure}[b]{0.3\linewidth}
 \centering 
 \centerline{\epsfig{figure=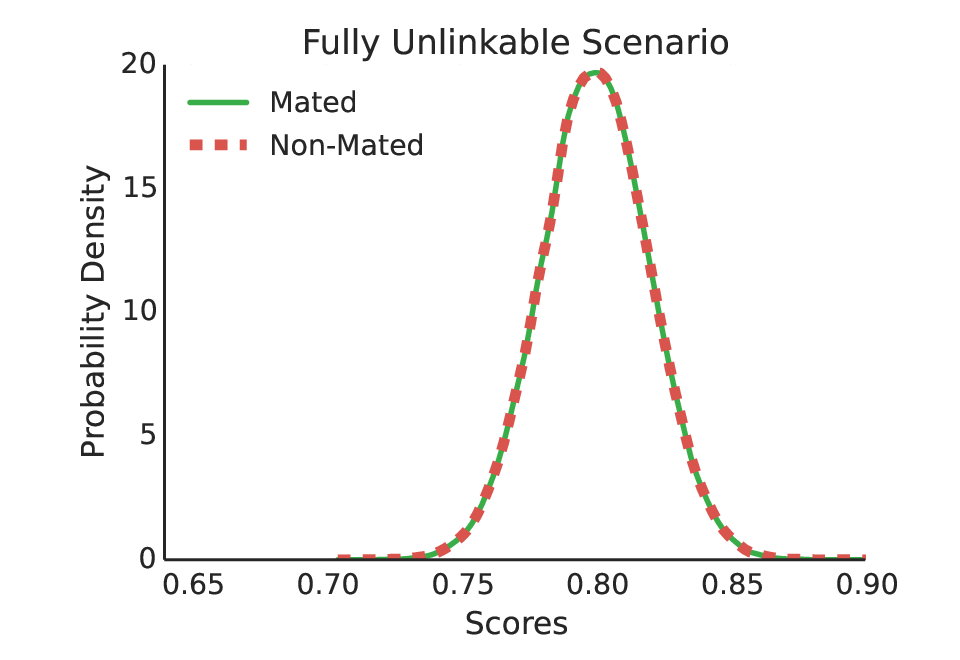,width=.99\linewidth}}\caption{}\label{fig:scenariosExample:full}
\end{subfigure}
\begin{subfigure}[b]{0.3\linewidth}
 \centering 
 \centerline{\epsfig{figure=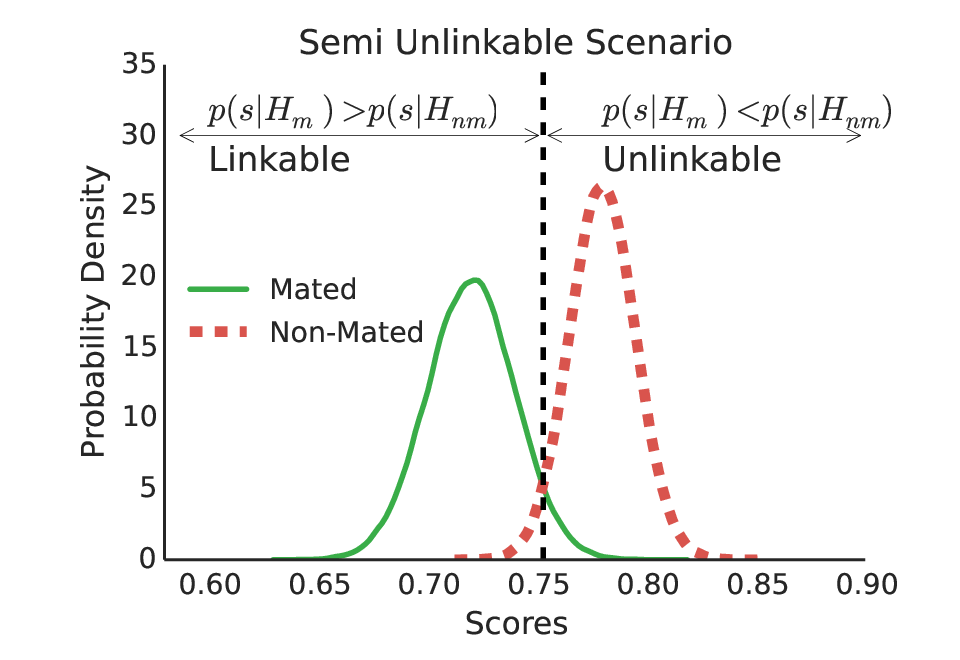,width=.99\linewidth}}\caption{}\label{fig:scenariosExample:semiLink}
\end{subfigure}
\begin{subfigure}[b]{0.3\linewidth}
 \centering 
 \centerline{\epsfig{figure=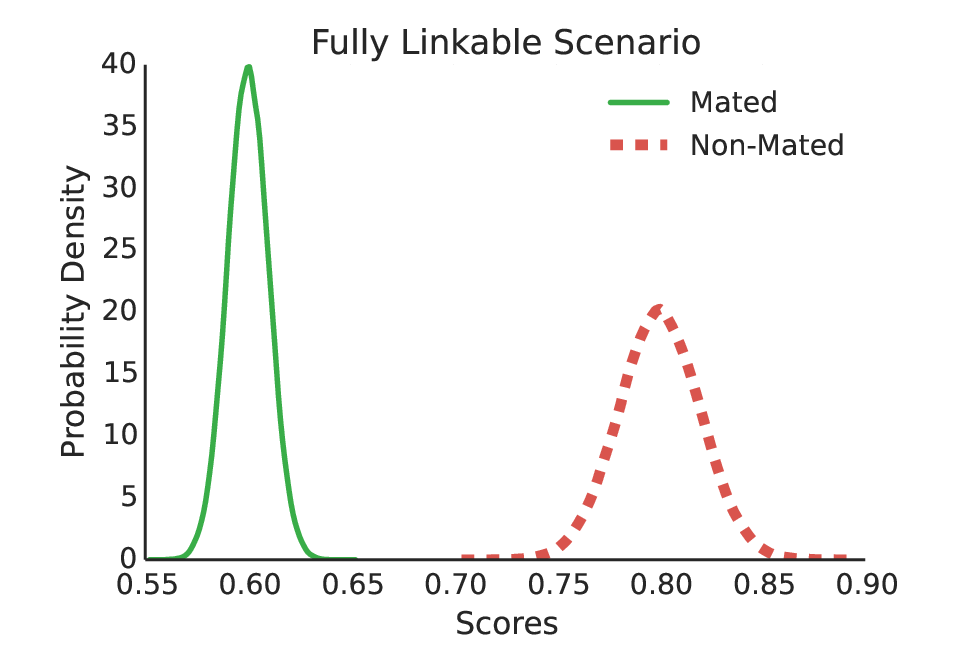,width=.99\linewidth}}\caption{}\label{fig:scenariosExample:fullLink}
\end{subfigure}
\caption{Examples of \textit{Mated samples} (solid green) and \textit{Non-mated samples} (dashed red) distibutions yielded by (a) fully unlinkable, (b) semi linkable and (c) fully linkable templates.} \label{fig:scenariosExample}\vspace{-0.5cm}
\end{figure*}

Now, to extend formality to the problem being addressed, some mathematical notations are introduced in this section. Let us define the following hypothesis:
\begin{align*}
&H_{m} = \{ \text{both templates belong to mated instances} \}
\\
&H_{nm} = \{ \text{both templates belong to non-mated instances} \}
\end{align*}

Based on those hypothesis, we can define two types of score distributions, where $s = \mathit{LS}\left( \mathbf{T}_1, \mathbf{T}_2 \right)$ is the linkage score between two templates, as defined in Fig.~\ref{fig:attack}:
\begin{itemize}

\item \textit{Mated samples} distribution: scores computed from templates extracted from samples of a single instance of the same subject and enrolled in different applications, using different keys:
\begin{equation}
\mathbf{T}_1 = \mathit{PIE}(\mathbf{M}_1,\mathbf{K}_1),\mathbf{T}_2 = \mathit{PIE}(\mathbf{M}_1, \mathbf{K}_2)
\end{equation} 
It represents the conditional probability of obtaining a score $s$ knowing that two templates come from mated instances, that is, $p\left( s | H_{m} \right)$.  

\item \textit{Non-mated samples} distribution: scores yielded by templates generated from samples of different instances and enrolled in different applications, using different keys: 
\begin{equation}
\mathbf{T}_1 = \mathit{PIE}(\mathbf{M}_1,\mathbf{K}_1),\mathbf{T}_2 = \mathit{PIE}(\mathbf{M}_2, \mathbf{K}_2)
\end{equation}
It represents the conditional probability of obtaining a score $s$ knowing that two templates come from non-mated instances, that is, $p\left( s | H_{nm} \right)$.
\end{itemize}

In contrast to \cite{Ferrara14a}, where three score distributions are considered, here only two distributions will be analysed as in \cite{wang2014cancelableFpConv}. The main reason is that the first distribution in \cite{Ferrara14a}, comprising linkage scores of exactly the same sample enrolled in different systems, is included here as part of the \textit{Mated samples} distribution (i.e., the mated samples distribution will represent both cases $i)$ and $ii)$). Furthermore, the score distribution related to case $i)$ in \cite{Ferrara14a} is a very unusual case in real scenarios as, in most cases, different enrolled samples will be acquired by each system. Nevertheless, the proposed framework is general and can also be directly applied to the case of having three score distributions.

\section{Measuring Linkability}
\label{sec:metric}



Based on the definition of linkability given in Sect.~\ref{sec:btp}, two templates are linkable to some degree if there is some method that allows us to determine that it is more likely that they come from the same instance than from different instances. Therefore, according to that definition, if a method allows us to determine that two templates are more likely to come from \textit{different} instances, those two templates are \textit{unlinkable}.


That is, if for a given score $s$ of the linkage function, $p\left( H_{m}  | s \right) > p\left( H_{nm} | s \right)$, then it is more likely that $\mathbf{T}_1$ and $\mathbf{T}_2$ belong to the same instance, and, therefore, the templates are linkable (to some degree). If, on the contrary, $p\left( H_{nm}  | s \right) \ge p\left( H_{m}  | s \right)$, then it is more likely that $\mathbf{T}_1$ and $\mathbf{T}_2$ belong to different instances. That is, for that score $s$, the linkage function would fail to link both templates. Therefore, it follows that, to have a fully unlinkable system for the whole domain of scores $s$ of the linkage function, there has to be a complete overlap between the aforementioned \textit{mated} and \textit{non-mated} score distributions.


Following the discussion presented to this point, we define two different measures for the linkability of biometric templates: 
\begin{itemize}
\item Local measure $\mathrm{D}_{\leftrightarrow} \left( s \right)$: it evaluates the linkability of the templates in a score-wise basis.

\item Global measure $\mathrm{D}_{\leftrightarrow}^{\mathit{sys}}$: it gives an overall measure of the linkability of the whole system, independent of the score domain of the system at hand, thereby allowing a benchmark among different systems.
\end{itemize} 

To better illustrate the relationship between the linkage scores and the proposed metrics, Fig.~\ref{fig:scenariosExample} represents generic \textit{Mated samples} distributions (in solid green) and \textit{Non-mated samples} distributions (in dashed red) for: (a) fully unlinkable, (b) semi-linkable and (b) fully linkable templates. As we observe,
\begin{itemize}
\item Under a \textit{fully unlinkable} scenario (Fig.~\ref{fig:scenariosExample:full}), both distributions are identical. In this case, the probability that, for a given score, the templates protect the same instances or different instances is the same and, therefore, the templates cannot be linked. 

\item Under a \textit{fully linkable} scenario (Fig.~\ref{fig:scenariosExample:fullLink}), the \textit{Mated samples} and \textit{Non-mated samples} distributions are fully separable. Therefore, given the linkage score between two templates, we can make a decision with almost all certainty for all the score domain on whether the templates that produced any of the scores stem from the same instance (i.e., $s \in [0.55, 0.65]$).

\item Under a \textit{semi linkable} scenario (Fig.~\ref{fig:scenariosExample:semiLink}), we observe that templates can be linked only for a subset of the scores: $s < 0.75$. For that subset, it is more likely that both templates stem from mated instances. For $s > 0.75$, templates are more likely to conceal different instances, and are not linkable.
\end{itemize}

In the following subsections, we describe how both the local and the global measures are computed and we discuss their main properties. 

\subsection{Local Measure $\mathrm{D}_{\leftrightarrow} \left( s \right)$: System Score-Wise Linkability}
\label{sec:metric:score}

$\mathrm{D}_{\leftrightarrow} \left( s \right) \in [0, 1]$ evaluates the linkability of a system for each \emph{specific linkage score} $s$. As such, this metric is appropriate to analyse within one system in which parts of the score domain it fails to provide unlinkability. If for a specific score $s_1$, a system yields $\mathrm{D}_{\leftrightarrow} \left( s_1 \right) = 1$, it means that, \emph{in case} the linkage function produced $s_1$, we would be able to link both templates $\mathbf{T}_1$ and $\mathbf{T}_2$ to the same instance with almost all certainty. On the other hand, $\mathrm{D}_{\leftrightarrow} \left( s_0 \right) = 0$ should be interpreted as full unlinkability for that particular score $s_0$. In other words, \emph{if} $s_0$ were produced by the linkage function, it would be more likely that both templates stemmed from different instances, hence failing to link them to a single data subject. All intermediate values of $\mathrm{D}_{\leftrightarrow} \left( s \right)$ between 0 and 1 report an increasing degree of linkability. 

As highlighted in Sect.~\ref{sec:btp} and Fig.~\ref{fig:scenariosExample:semiLink}, the key on the success of linking to templates lies on determining whether, given a score $s$, it is more likely that two templates stem from mated samples than from non-mated samples. Or, in other words, on whether $p\left( H_{m} | s \right) > p\left( H_{nm} | s\right)$ for a given score $s$. Therefore, such linkability can be accounted for in terms of the difference of conditional probabilities of each hypothesis $H_m$ and $H_{nm}$ for a given score $s$:
\begin{equation}
\mathrm{D}_\leftrightarrow\left(s\right) = p\left( H_{m} \vert s\right) - p\left( H_{nm} \vert s\right) \label{eq:Ddiff}
\end{equation}
However, these two conditional probabilities are unknown. As defined in Sect.~\ref{sec:btp}, what can be computed a priori, and is known for each biometric template protection system and linkage function, are the \textit{Mated} and \textit{Non-mated samples} distributions (i.e., $p\left(s|H_m\right)$ and $p\left(s|H_{mn}\right)$), that is, the probability of observing $s$ knowing that two templates, protected with different keys, belong to mated samples or to non-mated samples.




In the following we explain how $\mathrm{D}_{\leftrightarrow} \left( s \right) $ is computed based on the known distributions $p\left(s|H_m\right)$ and $p\left(s|H_{mn}\right)$. To do so, we start the computation from the likelihood ratio:
\begin{equation}\label{eq:LR}
LR \left( s \right) = \frac{ p\left( s \vert H_{m} \right) } { p\left( s \vert H_{nm} \right) } 
\end{equation}

Bayes theorem can be applied to Eq.~\ref{eq:LR} to obtain
\begin{equation}\label{eq:bayes}
LR \left( s \right) = \frac{ \frac{p\left( H_{m} \vert s\right) \cdot \cancel{p\left( s \right)}}{p\left( H_{m}\right)} } { \frac{p\left( H_{nm} \vert s\right) \cdot \cancel{p\left( s \right)}}{p\left( H_{nm}\right)} }  = \frac{ p\left( H_{m} \vert s\right) }{ p\left( H_{nm} \vert s\right) } \cdot \frac{p\left( H_{nm}\right)}{p\left( H_{m}\right)} 
\end{equation}

Therefore, the probabilities shown in Eq.~\ref{eq:Ddiff} are related as follows
\begin{equation}
\frac{p\left( H_{m} \vert s\right)}{p\left( H_{nm} \vert s\right)} = LR\left( s \right) \cdot \omega \label{eq:quotientH}
\end{equation}
where $\omega = p\left( H_{m}\right) / p\left( H_{nm}\right)$ denotes the ratio between the unknown prior probabilities of the \textit{Mated samples} and \textit{Non-mated samples} distributions. From this last equation, we have
\begin{equation}
p\left( H_{m} \vert s\right) = LR \left( s \right) \cdot  p\left( H_{nm} \vert s\right) \cdot \omega \label{eq:HmIni}
\end{equation}
Even if the values of those probability functions are unknown, $s$ must belong to any of those two distributions. As a consequence, 
\begin{equation}\label{eq:priors}
p\left( H_{nm} \vert s\right) = 1 - p\left( H_{m} \vert s \right)
\end{equation}
We can thus re-write Eq.~\ref{eq:HmIni} as
\begin{align}
p\left( H_{m} \vert s\right) &= LR \left( s \right) \cdot \left( 1 -  p\left( H_{m} \vert s\right)\right) \cdot \omega \Leftrightarrow
\\
p\left( H_{m} \vert s\right) &= \frac{LR \left( s \right) \cdot \omega}{1 + LR\left( s \right) \cdot \omega }\label{eq:Hm}
\end{align}

Now, as mentioned in Eq.~\ref{eq:Ddiff}, in order to define the local linkability measure $\mathrm{D}_\leftrightarrow\left(s\right)$ we are interested in the difference $p\left( H_{m} \vert s\right) - p\left( H_{nm} \vert s\right)$, which applying Eq.~\ref{eq:priors} yields:
\begin{equation}
\mathrm{D}_\leftrightarrow\left(s\right) = p\left( H_{m} \vert s\right) - p\left( H_{nm} \vert s\right) = 2\cdot p\left( H_{m} \vert s\right) - 1\label{eq:D1}
\end{equation}
Combining Eqs.~\ref{eq:Hm} and~\ref{eq:D1}, we can re-write the local linkability measure as
\begin{equation}
\mathrm{D}_\leftrightarrow\left(s\right) = 2\frac{LR \left( s \right) \cdot \omega}{1 + LR\left( s \right) \cdot \omega} - 1 \label{eq:D2}
\end{equation}

\begin{figure}[t]
\centering
\begin{subfigure}[b]{0.49\linewidth}
 \centering 
 \centerline{\epsfig{figure=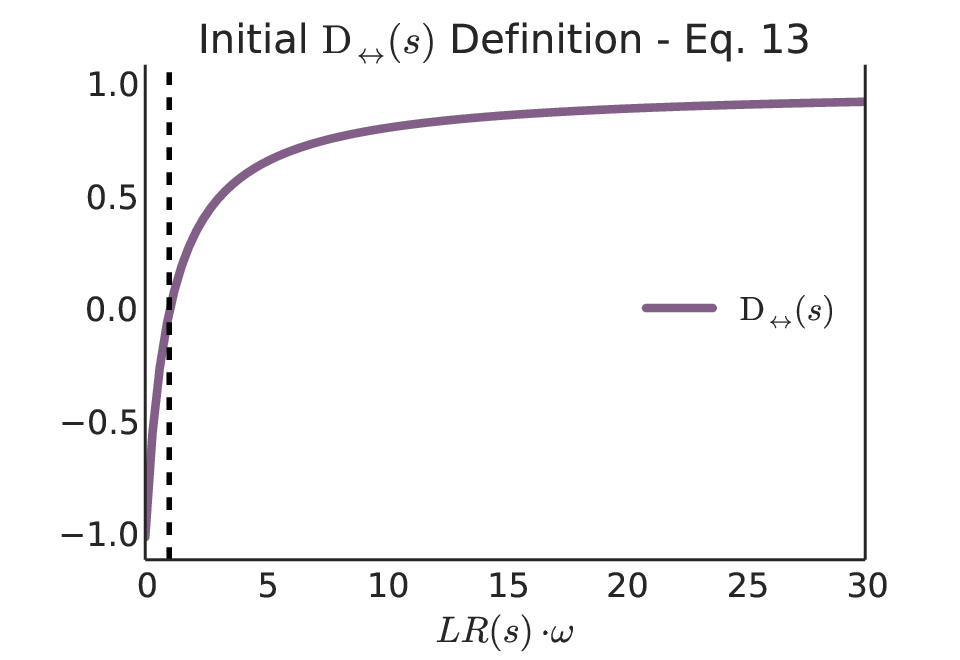,width=.99\linewidth}}\caption{}\label{fig:DlinkIni}
\end{subfigure}
\begin{subfigure}[b]{0.49\linewidth}
 \centering 
 \centerline{\epsfig{figure=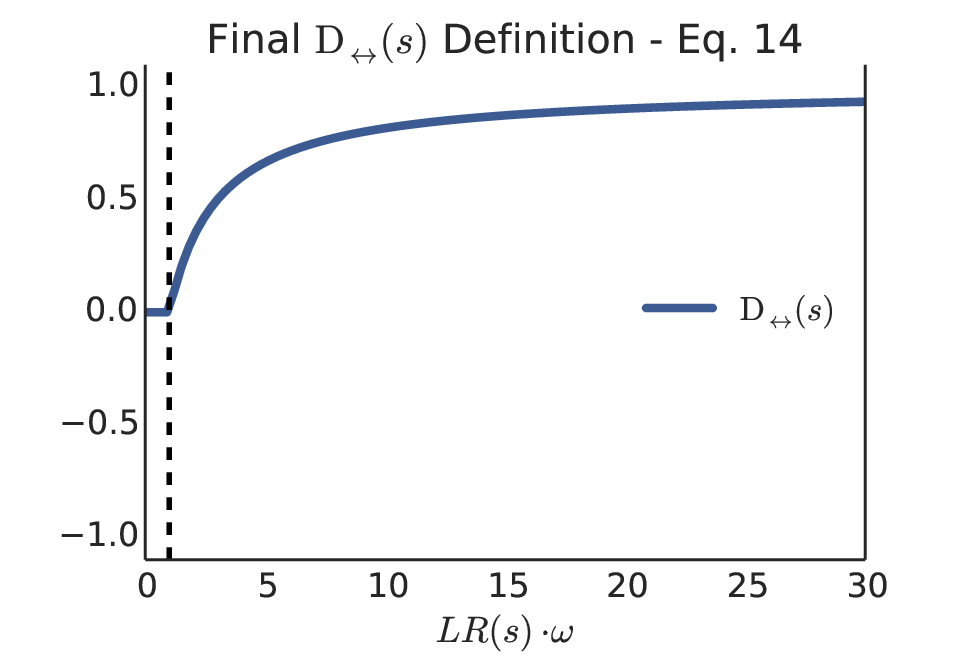,width=.99\linewidth}}\caption{}\label{fig:DlinkFinal}
\end{subfigure}
\caption{(a) Initial definition of $\mathrm{D}_{\leftrightarrow} \left( s \right)$, according to Eqs.~\ref{eq:Ddiff} and~\ref{eq:D2}, and (b) final definition of $\mathrm{D}_{\leftrightarrow} \left( s \right)$, according to Eq.~\ref{eq:unlink}, where $\mathrm{D}_{\leftrightarrow} \left( s \right) = 0$ for  $LR \left( s \right) \cdot \omega < 1$. The dashed black line represents $LR\left(s\right)  \cdot \omega = 1$, and thus $\mathrm{D}_{\leftrightarrow} \left( s \right) = 0$.} \label{fig:Dlink}\vspace{-0.4cm}
\end{figure}

This measure is depicted in Fig.~\ref{fig:DlinkIni}. As we may observe, this leads to two different scenarios, depending on the value of $LR\left( s \right) \cdot \omega$, separated by a dashed vertical line:
\begin{itemize}
\item If $LR\left( s \right) \cdot \omega \le1$ (i.e., left of the vertical line, where $p\left( H_{m} \vert s\right) \le p\left( H_{nm} \vert s\right)$), we can deduce, with some certainty, that both templates do \emph{not} belong to the same instance. As a consequence, we cannot link both templates to the same data subject. In that case, $\mathrm{D}_\leftrightarrow\left(s\right)$ yields a negative value, and hence the system is not linkable for that score. However, in order to have a single measure for unlinkable templates, a single value for this range of scores would be desired: $\mathrm{D}_\leftrightarrow\left(s\right) = 0 \quad \forall s \quad | \quad LR\left( s \right) \cdot \omega \le1$.


\item If $LR \left( s \right)  \cdot \omega > 1$ (i.e., right of the vertical line, where $p\left( H_{m} \vert s\right) > p\left( H_{nm} \vert s\right)$), we can state that it is more likely that both templates belong to mated instances, thereby making the templates somewhat linkable for those score values. In fact, in this case, the higher $LR\left( s \right)$, the more linkable the templates are. As a consequence, $\mathrm{D}_{\leftrightarrow} \left( s \right)$ yields an increasing value in $(0, 1]$, with higher values for more linkable templates (i.e., the higher $LR\left( s \right)$, the closer $\mathrm{D}_{\leftrightarrow} \left( s \right)$ is to 1).
\end{itemize} 

Keeping those remarks in mind, we define $\mathrm{D}_{\leftrightarrow} \left( s \right)  $ as a two-part function of $s$ in Fig.~\ref{fig:DlinkFinal}, depending on the value of the corresponding $LR\left( s\right) \cdot \omega$, as follows:
\begin{itemize}
\item $\mathrm{D}_{\leftrightarrow} \left( s \right) = 0$ for $s$ such that $LR\left( s\right)\cdot \omega \le 1$ (i.e., unlinkable score values). See Fig.~\ref{fig:DlinkFinal} for values $LR\left( s\right)\cdot \omega \le 1$ (i.e., values to the left of the vertical dashed line).

\item For the linkable score values (i.e., $LR\left( s\right) \cdot \omega> 1$), we define $\mathrm{D}_{\leftrightarrow} \left( s \right)$ as in Eq.~\ref{eq:D2}, hence defining a continuous function with  $\mathrm{D}_{\leftrightarrow} \left( s \right) = 0$ for $LR\left( s\right) \cdot \omega = 1$. 
\end{itemize}

Therefore, we finally have
\begin{equation}\label{eq:unlink}
\mathrm{D}_{\leftrightarrow}  \left( s \right) = 
\begin{cases}
0 &\text{if } LR\left(s\right) \cdot \omega \le 1
\\
2\frac{LR \left( s \right) \cdot \omega}{1 + LR\left( s \right) \cdot \omega} - 1&\text{if } LR\left(s\right) \cdot \omega > 1
\end{cases}
\end{equation}

This function $\mathrm{D}_{\leftrightarrow}  \left( s \right)$ presents some very interesting and desirable properties that make it specially well-behaved to be used as linkability metric (see Appendix~\ref{sec:annex:score} for the complete mathematical proof of these properties):
\begin{itemize}
\item Domain: $\mathrm{D}_{\leftrightarrow} \left( s \right)$ is defined over the whole score domain.

\item Continuity: $\mathrm{D}_{\leftrightarrow} \left( s \right)$ is continuous in the whole domain.

\item Range: $\mathrm{D}_{\leftrightarrow} \left( s \right)$ is bounded in $[0, 1]$.

\item Monotonicity: $\mathrm{D}_{\leftrightarrow} \left( s \right)$ is a monotonically increasing function.
\end{itemize} 

It should be noted that in the definition we have a variable $s$ (i.e., the linkage score) and a constant value $\omega$ (i.e., the quotient between the prior probabilities). For a given system, if the prior probabilities are available, those statistics should be used to compute $\omega$. Otherwise, and in order not to bias the analysis towards one type of comparison, we can assume that both mated and non-mated comparison attempts are equally probable (i.e., worst-case scenario in the unlinkability analysis, see below). In other words, we assume that $p\left(H_m\right) = p\left( H_{nm}\right)$, and thus set $\omega = 1$.

Regarding those prior probabilities, it should be highlighted that they are different from the mated and non-mated probabilities derived from normal verification attempts, where we can assume $ p\left(H_m\right) \gg p\left( H_{nm}\right) $. For the particular case of the unlinkability analysis and linkage scores, comparisons are carried out between a given template, enrolled in application A, and all templates enrolled in application B. Therefore, assuming that $N$ different subjects are enrolled in database B, the probability of obtaining a mated and a non-mated score are, respectively:
\begin{eqnarray}
p\left(H_m\right) &=& 1 / N
\\
p\left(H_{nm}\right) &=& \left( N - 1 \right) / N
\end{eqnarray}
As a consequence, for all $N \ge 2$, $ p\left(H_m\right) \le p\left( H_{nm}\right) $, being equal only for $N = 2$. That leads to values $\omega \le 1$, being $\omega = 1$ the worst-case scenario for the unlinkability evaluation, where only two subjects are enrolled in system B. For a more in depth analysis of this parameter, a specific experiment is performed in Sect.~\ref{sec:exp:omega}.

\subsection{Global Measure $\mathrm{D}_{\leftrightarrow}^{\mathit{sys}}$: System Overall Linkability}
\label{sec:metric:global}

As described previously, it is also useful to have an estimation of the \emph{unlinkability of the whole system} (and not just on a score-wise basis as is the case of $\mathrm{D}_{\leftrightarrow}\left( s\right)$), which may allow a fairer benchmark of the unlinkability level of two or more systems. 

For this purpose, we introduce the global metric $\mathrm{D}_{\leftrightarrow}^{\mathit{sys}} \in [0, 1]$, which gives an estimation of the global linkability of a system, \emph{independently} of the score. This way, if a system has $\mathrm{D}_{\leftrightarrow}^{\mathit{sys}} = 1$ (i.e., case in which both the \textit{Mated samples} and \textit{Non-mated samples} distributions have no overlap, as shown in Fig.~\ref{fig:scenariosExample:fullLink}), it means that it is fully linkable for all the scores of the \textit{Mated samples} distribution domain (i.e., where $\mathrm{D}_\leftrightarrow\left(s\right) = 1$). That is, if we evaluate the linkage function on two protected templates $\mathbf{T}_1$ and $\mathbf{T}_2$, we can decide (with almost all certainty) whether they conceal or not the same instance. Similarly, $\mathrm{D}_{\leftrightarrow}^{\mathit{sys}} = 0$ (i.e., Fig.~\ref{fig:scenariosExample:full}, where both score distributions totally overlap) means that the system is fully unlinkable for the whole score domain. That is, independently of the score produced by the linkage function, it is equally probable that the two templates stem from the same instance ($H_{m}$) than from different instances ($H_{nm}$). All intermediate values of $\mathrm{D}_{\leftrightarrow}^{\mathit{sys}} $ between 0 and 1 report a decreasing degree of unlinkability (i.e., increasing degree of linkability).


Therefore, we are interested on measuring how likely it is to get a score stemming from the \textit{Mated samples} distribution. This can be achieved computing the difference $p\left( H_m \cap s\right) - p\left( H_{nm} \cap s\right)$ and integrating it over the whole score domain:
\begin{equation}
\begin{split}
\int &p\left( H_m \cap s\right) - p\left( H_{nm} \cap s\right)\mathrm{d}s = 
\\
\int &p\left( s\right) \cdot \left( p\left( H_m | s\right) - p\left( H_{nm} | s \right)\right)\mathrm{d}s =
\\
 p \left( H_m \right) & \int p\left( s | H_m\right) \cdot \left( p\left( H_m | s\right) - p\left( H_{nm} | s \right)\right)\mathrm{d}s +
\\
p\left( H_{nm} \right) &\int p\left( s | H_{nm} \right) \cdot \left( p\left( H_m | s\right) - p\left( H_{nm} | s \right)\right)\mathrm{d}s
\end{split} \label{eq:DglobalIni}
\end{equation}
The second equality holds since $p\left( s \right) = p\left( s | H_m\right) \cdot p\left( H_m \right) + p\left( s | H_{nm} \right) \cdot p\left( H_{nm} \right)$. 

Regarding the success on linking templates, we are only interested in the first addend, that is, in the probabilities stemming from the \textit{Mated samples} distribution. In addition, as in the definition of the local linkability measure $\mathrm{D}_{\leftrightarrow}\left( s \right)$, two templates can be linked only if $p\left(H_m | s \right) > p\left(H_{nm} | s \right)$. Therefore, we define $\mathrm{D}_{\leftrightarrow}^{\mathit{sys}}$ as
\begin{equation}\label{eq:unlinkSystemIni}
\mathrm{D}_{\leftrightarrow}^{\mathit{sys}}= \int\displaylimits_{\begin{split}p&\left(H_m | s \right) > \\ &p\left(H_{nm} | s \right)\end{split}} p\left( s | H_m\right) \cdot \left( p\left( H_m | s\right) - p\left( H_{nm} | s \right)\right)\mathrm{d}s 
\end{equation}
where the first addend in Eq.~\ref{eq:DglobalIni} has been normalised by $p\left( H_m \right)$ in order to obtain a measure independent of the prior probability. 

Keeping that in mind, we can finally re-write Eq.~\ref{eq:unlinkSystemIni} in terms of $\mathrm{D}_{\leftrightarrow}  \left( s \right) $ and $p\left( s \vert H_{m} \right)$, hence defining $\mathrm{D}_{\leftrightarrow}^{\mathit{sys}}$ as:
\begin{equation}\label{eq:unlinkSystem}
\mathrm{D}_{\leftrightarrow}^{\mathit{sys}}= \int p\left( s \vert H_{m} \right) \cdot \mathrm{D}_{\leftrightarrow}  \left( s \right)  \mathrm{d}s
\end{equation}
This way, the final value of $\mathrm{D}_{\leftrightarrow}^{\mathit{sys}}$ depends on: $i)$ the domain of scores where the system is linkable (determined by $\mathrm{D}_\leftrightarrow\left(s\right)$); $ii)$ how linkable the system is in that domain of scores (given by $\mathrm{D}_\leftrightarrow\left(s\right)$); and $iii)$ how probable it is that such scores are produced (given by $p\left(s | H_m\right)$). Therefore, this new global measure assigns different levels of linkability to intermediate scenarios, not fully unlinkable or fully linkable.

Derived from the properties of the well-behaved $\mathrm{D}_\leftrightarrow\left( s\right)$ function, the global metric $\mathrm{D}_{\leftrightarrow}^{\mathit{sys}}$ also presents some desirable properties (see Appendix~\ref{sec:annex:global} for the complete mathematical proof):
\begin{itemize}
\item $\mathrm{D}_{\leftrightarrow}^\mathit{sys}$ is properly defined.

\item Range: $\mathrm{D}_{\leftrightarrow}^\mathit{sys}$ is bounded in $[0,1]$.
\end{itemize}


\subsection{Local and Global Metrics: Discussion}
\label{sec:metric:discussion}

Since the mated and non-mated score distributions are statistical distributions, their integral between $-\infty$ and $+\infty$ is exactly one. As a consequence, if there is a range of scores $\{s_{\mathrm{linkable}}\}$ where $p\left(s | H_m\right) > p\left(s | H_{nm}\right)$, there must be a range of scores $\{s_{\mathrm{unlinkable}}\}$ where $p\left(s | H_{nm}\right) > p\left(s | H_{m}\right)$. Therefore, given the definition of unlinkability presented in Section~\ref{sec:btp}, if two templates $\mathbf{T}_1$ and $\mathbf{T}_2$ are compared using a linkage function and they produce a score $s_0$ that falls in $s_0 \in \{s_{\mathrm{unlinkable}}\}$, the two templates are unlinkable. In other words, for that score $s_0$ the system is unlinkable, and hence $\mathrm{D}_\leftrightarrow \left(s_0\right)=0$. On the other hand, by obtaining that score $s_0$ we have learned something about the system: there must be some other score $s_1 \in \{s_{\mathrm{linkable}}\}$ where the system is linkable, because, as mentioned above, we are dealing with statistical distributions. 

In summary, if we obtain $s_0$ where $p\left(s_0 | H_{nm}\right) > p\left(s_0 | H_{m}\right)$, we know that: $i)$ the system is unlinkable for $s_0$, $\mathrm{D}_\leftrightarrow \left(s_0\right)=0$; $ii)$ the system must be linkable for some $s_1$, whose value we don't know and for which $\mathrm{D}_\leftrightarrow \left(s_1\right) \ne 0$; and $iii)$ as a consequence, the global measure $\mathrm{D}_\leftrightarrow^\mathit{sys} \ne 0$. Therefore, by obtaining a score in the unlinkable range of scores $s_0$, we actually learn that the system, as a whole, is linkable at some point, and this will be reflected by $\mathrm{D}_\leftrightarrow^\mathit{sys} $. However, this does not mean that $\mathrm{D}_\leftrightarrow \left(s_0\right)$ should be different from 0, because, given the unlinkability definition, for that specific score, the system is not linkable (i.e., we cannot link the two templates to the same instance, on the contrary, we can link them to different data subjects). 

\section{Proposed Linkability Evaluation Protocol}
\label{sec:protocol}

After presenting and showing the complementarity of both unlinkability measures (i.e., local and global), we propose here a general protocol for the evaluation of the unlinkability of a particular biometric template protection system. 

\begin{figure*}[t]
\centering
\begin{subfigure}[b]{0.32\linewidth}
 \centering 
 \centerline{\epsfig{figure=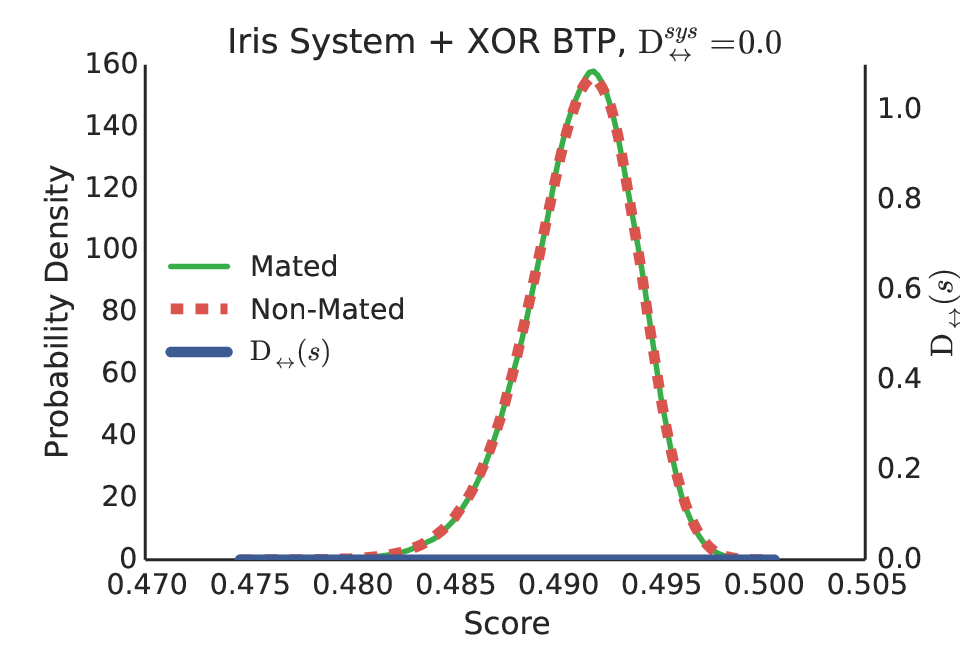,width=.99\linewidth}}\caption{}\label{fig:scores:irisBF}
\end{subfigure}
\begin{subfigure}[b]{0.32\linewidth}
 \centering 
 \centerline{\epsfig{figure=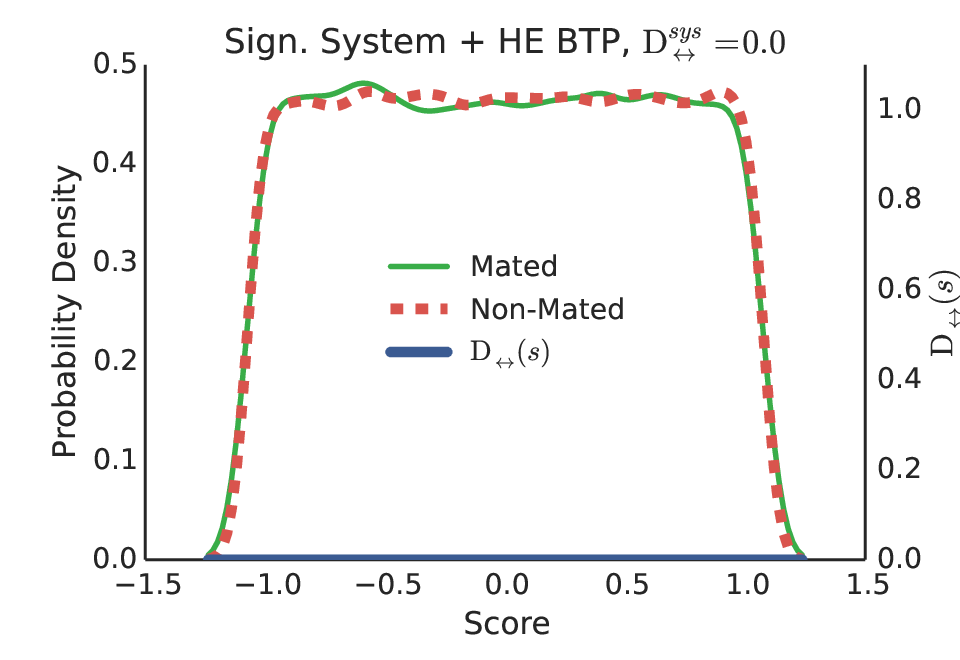,width=.99\linewidth}}\caption{}\label{fig:scores:signHE}
\end{subfigure}
\begin{subfigure}[b]{0.32\linewidth}
 \centering 
 \centerline{\epsfig{figure=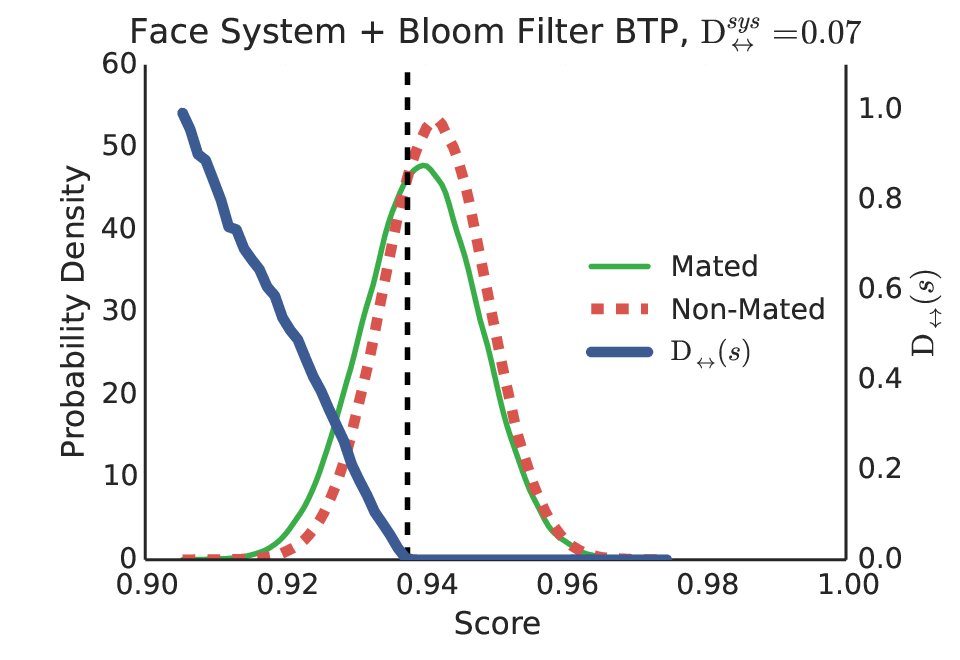,width=.99\linewidth}}\caption{}\label{fig:scoresFace:faceBF}
\end{subfigure}
\caption{Examples of unlinkability analysis for three different BTP systems. The dashed vertical lines represent $LR\left( s\right) = 1$.} \label{fig:scores}\vspace{-0.4cm}
\end{figure*}

First, it should be noted that, in practice, linkability is defined as the ability to link templates across different applications (i.e., stored in databases used by different applications). Templates stored in different databases are protected using different keys. As a consequence, in order to evaluate the linkability of templates protected with a specific algorithm, several different databases need to be created, containing templates extracted from the same biometric samples and using a different key for each database. With this in mind, the proposed protocol to evaluate the linkability of templates would run as follows:
\begin{enumerate}
\item Generate $K$ databases of protected templates each of them using a different key. It is recommended that $K>5$.

\item Compute the \textit{Mated samples} score distribution for the selected linkage function, \textit{across the $K$ databases} generated in step 1. As in any recognition accuracy evaluation of a traditional biometric system, the larger the database, the more statistically significant the results of the evaluation are. Please see the end of this section for some relevant examples on how to conduct a biometric accuracy evaluation.

\item Compute the \textit{Non-Mated samples} score distribution for the selected linkage function, \textit{across the $K$ databases} generated in step 1. Analogous considerations as those described for the \textit{Mated samples} score distribution should be taken into account.



\item If the prior probabilities $p\left(H_m\right)$ and $p\left( H_{nm}\right)$ are available, use them to compute $\omega$. Otherwise, we can assume that $p\left(H_m\right) = p\left( H_{nm}\right)$, and thus set $\omega = 1$.

\item Compute $\mathrm{D}_\leftrightarrow\left(s \right)$.

\item Compute $\mathrm{D}_\leftrightarrow^\mathit{sys}$.

\item Report $\mathrm{D}_\leftrightarrow\left(s \right)$ plots, together with the \textit{Mated samples} and \textit{Non-Mated samples} distributions, and the corresponding global linkability values $\mathrm{D}_\leftrightarrow^\mathit{sys}$, as in the examples analysed Sect.~\ref{sec:results} (Figs.~\ref{fig:scores} and~\ref{fig:scoresAttack}).

\item Analyse the plots and $\mathrm{D}_\leftrightarrow^\mathit{sys}$ values: where does $\mathrm{D}_\leftrightarrow\left(s \right)$ reach a maximum? Is it close to one? Where is $LR\left( s\right) \cdot \omega= 1$? What is the global value $\mathrm{D}_\leftrightarrow^\mathit{sys}$?
\end{enumerate}

It should be noted that the aforementioned steps should be repeated for each linkage function considered, always following the same score computation protocol. This way, the more functions considered, the more significant and thorough the evaluation is. In particular, if $F$ different functions are used in parallel to link the templates, $F$ scores $s_1 = \mathit{LS}_1\left(\mathbf{T}_1, \mathbf{T}_2\right), \dots, s_F = \mathit{LS}_F\left(\mathbf{T}_1, \mathbf{T}_2\right)$ will be obtained for a single pair of templates. Those scores will lead to the corresponding global linkability values $\mathrm{D}_{\leftrightarrow, 1}^\mathit{sys}, \dots, \mathrm{D}_{\leftrightarrow, F}^\mathit{sys}$. As a consequence, the system will be at least as vulnerable as the most challenging function considered: $\mathrm{D}_{\leftrightarrow}^\mathit{sys} = \max_f{ \{\mathrm{D}_{\leftrightarrow, f}^\mathit{sys}\}}$.

In addition, and in order to make the evaluation reproducible, the following information should be reported for each linkage function:
\begin{itemize}
\item The database and particular protocol followed for computing the \textit{Mated samples} and \textit{Non-Mated samples} distributions. If two or more systems are compared, the same database and protocol should be followed in order to allow for a fair benchmark. A good example of how to define such protocol may be found in \cite{ferrara12MCCnonInv,mansfield02best}.

\item Details on how the linkage function is computed (e.g.., input, knowledge required).

\item Define the type of information the attacker has access to. As in any other security evaluation, it is important to set the adversary model considered in the evaluation. It should be noticed that, depending on the security model considered, the level of linkability of a system may vary.
\end{itemize}

In summary, it is important to notice that the level of linkability of a system depends on: $i)$ the type of linkage function considered, and $ii)$ the security model assumed for the evaluation. 

The linkability protocol defined above is general in the sense that it can be applied to any \textit{Lebesgue integrable} linkage function and any security model, therefore providing a general framework to benchmark linkability across systems (or for one single system under different linkage functions and / or adversary models). We believe that this assumption on the nature of the scores is not very restrictive, since a majority of the usual linkage functions used in biometric recognition produce continuous normalizable scores (e.g., distance-based functions). Nonetheless, the metrics could eventually be applied to other non-continuous linkage functions by mapping them to a distance-based function.

Finally, while the computation of mated and non-mated sets of scores is part of the linkability protocol described in this section, the description of how to perform such accuracy evaluation falls out of the scope of the present article. However, the interested reader can find plenty of works in the literature addressing this problem, being two good examples \cite{ferrara12MCCnonInv,mansfield02best}.

In order to facilitate the use of the present framework and allow reproducibility of the article, a Python implementation of the metrics will be made available through the da/sec website\footnote{\url{https://dasec.h-da.de/research/biometrics/unlinkability/}} and the da/sec Github account\footnote{\url{https://github.com/dasec/unlinkability-metric}}.

\section{Analysing the Unlinkability of BTP Systems}
\label{sec:results}

In this section, following the protocol proposed in Sect.~\ref{sec:protocol}, we apply the metrics described in Sect.~\ref{sec:metric} to evaluate the unlinkability of four previously proposed BTP schemes. In addition to that analysis, the impact of using different values for $\omega$ is studied in Sect.~\ref{sec:exp:omega} and a comparison with other metrics proposed in the literature is carried out in Sect.~\ref{sec:exp:comp}.

\subsection{Experimental Setup}
\label{sec:exp:setup}

To show the generality of the proposed framework, four systems based on different biometric characteristics, using different features and comparators, and protected with different BTP algorithms, will be considered.

\textbf{Iris verification + random XOR protection}: as unprotected system we use a particular implementation within the publicly available University of Salzburg Iris Toolkit v1.0\footnote{\url{http://www.wavelab.at/sources/}}\cite{Uhl12a} of the Log-Gabor based algorithm proposed by Masek \cite{masek03irisSystem}. Dissimiliarity scores are computed in terms of the Hamming Distance. Iris-codes are then protected by XORing them with random binary strings \cite{zuo08verticalShiftIris}.

\textbf{On-line signature verification + Homomorphic Encryption protection}: as unprotected system, a state-of-the-art approach based on global features has been chosen \cite{marcos13dtw}. From a set of 100 global features extracted from the x and y coordinates, the best 40 normalized features according to \cite{galbally08GlobalFeatSelection} are selected to form the final template. Dissimilarity scores are computed based on the Euclidean distance. Then, templates are protected with Homomorphic Encryption \cite{marta16HEGLobalFeats}.

\textbf{Face verification + Bloom filter protection}: in the unprotected domain, the Log-Gabor Binary Pattern Histograms Sequences algorithm proposed in \cite{zhang05LGBPHS} is used. In particular, experiments are run on a publicly available implementation within the FaceRecLib\footnote{\url{https://pypi.python.org/pypi/facereclib}} \cite{gunther12facereclib} and the Bob Toolbox. Dissimilarity scores are computed based on histograms intersections. Templates are then protected using Bloom filters\footnote{\url{https://github.com/dasec/face-bf-btp}} \cite{marta16BFsUnlink}.

\textbf{Fingervein verification + block re-mapping protection}: as unprotected system we use the maximum curvature method presented in \cite{miura07MCPfingervein} to extract the connected vein pattern\footnote{\url{https://de.mathworks.com/matlabcentral/fileexchange/35716-miura-et-al-vein-extraction-methods}}. Similarity scores are computed in terms of the cross-correlation of rotated templates. Finally, block re-mapping \cite{ratha01securityPrivacy} is applied to protect the templates.

Following the recommendations given in the linkability evaluation protocol presented in Sect.~\ref{sec:protocol}, in the experiment we have considered $K = 10$ different keys to protect the templates. This simulates a case in the real world where the same subjects are enrolled in ten different applications and an attacker tries to link the templates in the ten databases to each other. Subsequently, the \textit{Mated samples} and \textit{Non-Mated samples} distributions are computed from scores stemming from templates stored in different databases (i.e., protected with different keys).

Keeping those remarks in mind, the iris, signature and face systems have been analysed over the corresponding sub-corpora of the multimodal BioSecure database \cite{biosecure09PAMI}, which comprises data of 210 different subjects. The iris and face sub-corpora include four samples of each eye and face, respectively (in the present study only the left eye sample is considered). On the other hand, the on-line signature sub-corpus comprises sixteen samples of each subject. In particular, for the \textit{Mated samples} distribution, the following scores are computed: $i)$ for the iris- and face-based systems, all possible mated comparisons where the reference sample is protected with a different key than the probe sample (i.e., 56,700 scores), and $ii)$ for the signature-based systems, the first four signatures are used for enrolment and the twelve remaining ones are used to compute the scores (i.e., 113,400 scores). Then, for the \textit{Non-Mated samples} distribution, in both cases the scores are obtained comparing one sample of the iris, face or the signature stored reference, with the first sample of the remaining subjects, protected with different keys (i.e., 987,525 scores all both characteristics).

The fingervein system used for the comparison with the state-of-the-art in unlinkability analysis is evaluated on the UTFVP database\footnote{\url{http://www.sas.el.utwente.nl/home/datasets}} \cite{UTFVPDB}. The database comprises data from 60 different subjects, from whom the vascular pattern of the index, ring and middle finger of both hands was collected twice at each of the two acquisition sessions ($60 \times 6 \times 4 = 1,440$ fingervein samples). In the experiments, for the \textit{Mated samples} distribution all possible scores are computed (i.e., 2,160 scores). Then, for the \textit{Non-Mated samples} distribution, the scores are obtained comparing the enrolled fingervein with the first sample of the remaining instances, protected with different keys (i.e., 64,520 scores).

Regarding implementation details, LRs are computed in a point-wise fashion. In addition, since we have no prior evidence about the unknown prior probabilities of the \textit{Mated samples} and \textit{Non-mated samples} distributions to estimate the most appropriate value for $\omega$, we will assume that $p\left( H_{m}\right) = p\left( H_{nm}\right)$, and hence $\omega = 1$.

Finally, the experimental evaluation comprises four stages: $i)$ the iris, signature and face systems are analysed in Sect.~\ref{sec:results:PIC}, utilising the dissimilarity score of the original $\mathit{PIC}$ (Pseudonymous Identifier Comparator) as linkage function, $ii)$ then further linkage functions trying to exploit some extra-knowledge or vulnerability of the original $\mathit{PIC}$ are studied in Sect.~\ref{sec:results:attacks} for the face-based system, $iii)$ the impact of different values of $\omega$ on the metrics is subsequently analysed in Sect.~\ref{sec:exp:omega} for both the facial and iris based templates, and $iv)$ the fingervein system is evaluated with both the newly proposed metric and other general metrics described in Sect.~\ref{sec:related} in order to show the advantages of the presented framework.

\begin{figure*}[t]
\centering
\begin{subfigure}[b]{0.32\linewidth}
 \centering 
 \centerline{\epsfig{figure=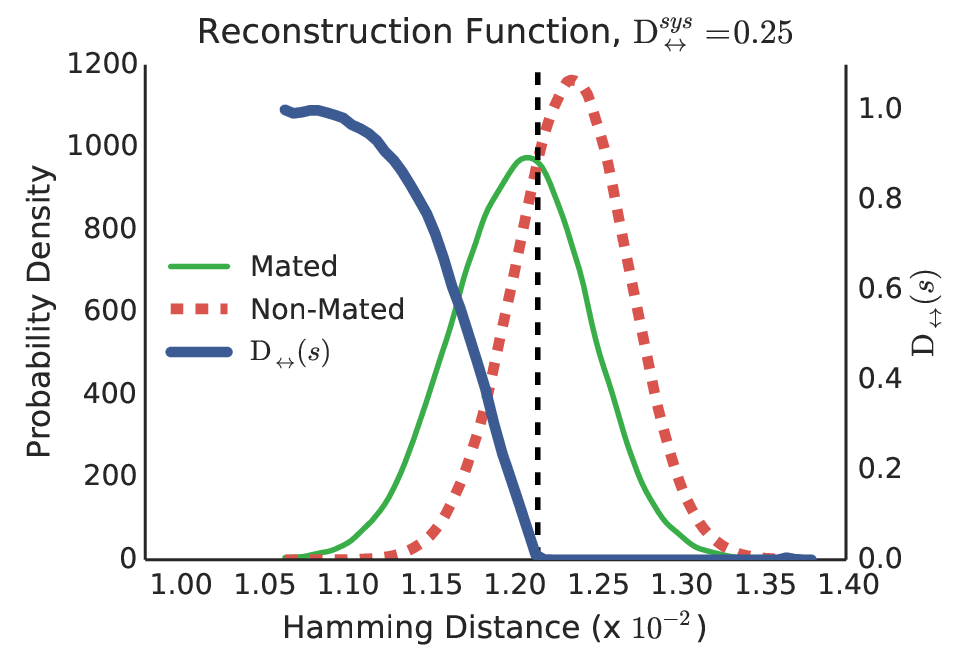,width=.99\linewidth}}\caption{}\label{fig:scoresAttack:rec}
\end{subfigure}
\begin{subfigure}[b]{0.32\linewidth}
 \centering 
 \centerline{\epsfig{figure=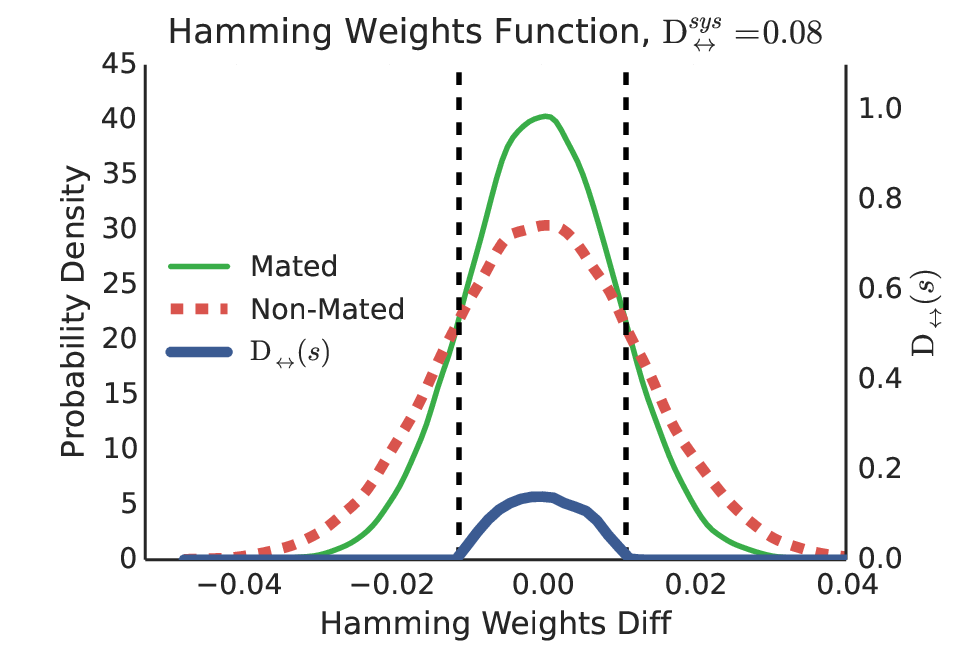,width=.99\linewidth}}\caption{}\label{fig:scoresAttack:hw}
\end{subfigure}
\begin{subfigure}[b]{0.32\linewidth}
 \centering 
 \centerline{\epsfig{figure=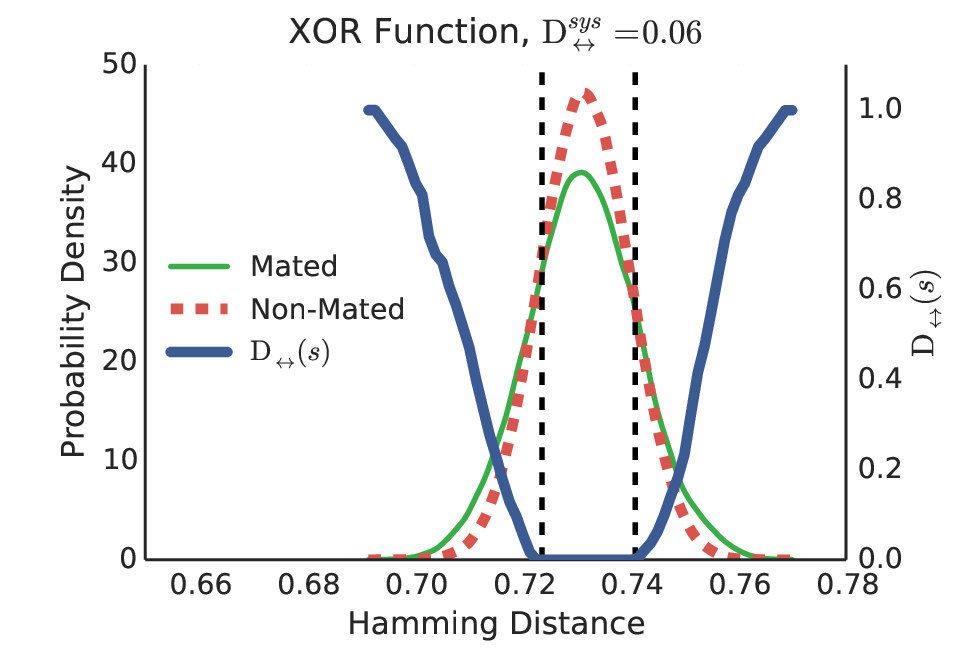,width=.99\linewidth}}\caption{}\label{fig:scoresAttack:xor}
\end{subfigure}
\caption{Unlinkability analysis of the facial BTP scheme under three linkage functions different from the original $\mathit{PIC}$.} \label{fig:scoresAttack}\vspace{-0.4cm}
\end{figure*}

\subsection{First Linkage Function: Systems' $\mathit{PIC}$ Scores}
\label{sec:results:PIC}

The first set of linkage functions analysed are the original $\mathit{PIC}$s described for each of the four protected systems considered \cite{zuo08verticalShiftIris,marta16HEGLobalFeats,marta16BFsUnlink,BRatha01a}. As such, in this case we only assume knowledge of the dissimilarity score computed by each scheme: $s = LS \left( \mathbf{T}_1, \mathbf{T}_2\right) = \mathit{PIC} \left(\mathbf{T}_1, \mathbf{T}_2\right)$. No further knowledge is assumed. Therefore, this is the trivial \lq\lq zero-effort'' linkage score, where only $\mathit{PIC}$ based comparisons between protected templates are considered.

The corresponding \textit{Mated samples} (solid green) and \textit{Non-mated samples} (dashed red) distributions are depicted in Fig.~\ref{fig:scores}: iriscodes protected with an XOR with a random string in Fig.~\ref{fig:scores:irisBF}, on-line signature protected with Homomorphic Encryption in Fig.~\ref{fig:scores:signHE}, and face protected with Bloom filters in Fig.~\ref{fig:scoresFace:faceBF}. In all cases, the proposed score-wise linkability measure $\mathrm{D}_{\leftrightarrow}  \left( s \right)$ is depicted in blue, and the global measure $\mathrm{D}_{\leftrightarrow}^{\mathit{sys}}$ is shown in the corresponding figure title. 


We may observe in Figs.~\ref{fig:scores:irisBF} and~\ref{fig:scores:signHE} that both types of protected templates are robust to linkage functions based on the dissimilarity score of the \textit{PIC}. As it may be seen, the \textit{Mated samples} and \textit{Non-mated samples} distributions completely overlap, leading to $\mathrm{D}_{\leftrightarrow}  \left( s \right) = 0$ for the whole domain of scores. Accordingly, $\mathrm{D}_{\leftrightarrow}^{\mathit{sys}} = 0$, as it corresponds to fully unlinkable templates. On the other hand, face protected templates (Fig.~\ref{fig:scoresFace:faceBF}) are still slightly linkable, showing a global linkability measure of $\mathrm{D}_{\leftrightarrow}^{\mathit{sys}} = 0.07$. This is due to the fact that, for dissimilarity scores $s < 0.94$, it is more likely that templates stem from mated instances. However, since the probability of obtaining such scores is very low ($p\left(s | H_{m}\right) < 0.005$), the system is almost fully unlinkable - hence the low value for $\mathrm{D}_{\leftrightarrow}^{\mathit{sys}}$.


Therefore, in these template protection systems, the proposed metrics allow us to conclude, based on objective values, that the BTP approaches implemented to secure the templates serve their purpose and provide full or a very high degree of unlinkability.

\subsection{Further Linkage Functions Different from the $\mathit{PIC}$}
\label{sec:results:attacks}

As it was highlighted in Sect.~\ref{sec:protocol}, a complete unlinkability analysis should also include the evaluation of the robustness of the system to specifically designed linkage functions which exploit particular vulnerabilities of BTP approaches \cite{simoens12BTPmetrics}. To that end, the aforementioned distributions should be estimated not only for the $\mathit{PIC}$ score of the system, but also for other more sophisticated functions and their corresponding linkage scores. Accordingly, we now analyse the vulnerabilities of the protected face verification system described above to three additional linkage functions. In all cases, the same protocol for the score computation described for the face case study in Sect.~\ref{sec:exp:setup} has been followed, and analogous distributions for the corresponding linkage scores are depicted in Fig.~\ref{fig:scoresAttack}. As it may be oserved, the score distributions are different from those obtained for the first linkage function based on the $\mathit{PIC}$ score (see Fig.~\ref{fig:scoresFace:faceBF}), since different information is exploited to discriminate between mated and non-mated comparisons. A brief description of the linkage functions provided below, and for more details on the linkage functions, the reader is referred to \cite{bringer15secAnalysisBF,hermans14bloomDoom,marta16BFsUnlink}.




\textbf{Reconstruction function} (Fig.~\ref{fig:scoresAttack:rec}): a methodology for the reconstruction of iriscodes given their corresponding Bloom filter based templates is proposed in \cite{bringer15secAnalysisBF}, which can be also used to link templates. In this case, the linkage function will operate in a two-stage manner: $i)$ reconstruct each unprotected template ($\mathbf{T}_1^{\mathrm{rec}}$, $\mathbf{T}_2^{\mathrm{rec}}$) from the protected counterpart ($\mathbf{T}_1$, $\mathbf{T}_2$), and $ii)$ compute the Hamming distance between the reconstructed templates: $s = LS \left( \mathbf{T}_1, \mathbf{T}_2\right) = \mathit{HD}\left(\mathbf{T}_1^{\mathrm{rec}}, \mathbf{T}_2^{\mathrm{rec}}\right)$. Therefore, for the reconstruction step, knowledge of the secret key of the system is assumed. 
 
As we may observe in Fig.~\ref{fig:scoresAttack:rec}, for half of the scores ($s < 1.2\times 10^{-2}$), it is more likely that the templates which yielded $s$ conceal the same instance. Which, in turn, leads to a successful linkage of the templates (i.e., $\mathrm{D}_\leftrightarrow\left( s\right) > 0$). Moreover, for a wide part of those scores ($s < 1.15\times 10^{-2}$), we can assume that with almost all certainty the two compared templates belong to the same instance, since $p\left(s | H_{nm}\right) = 0$ and $p\left(s | H_{m}\right) > 0$. Consequently, $\mathrm{D}_\leftrightarrow\left( s\right) = 1$ in that range. The vulnerabilities to this linkage function are also reflected in the global linkability measure, $\mathrm{D}_\leftrightarrow^\mathit{sys} = 0.25$, which is a value higher than that of the $\mathit{PIC}$ score ($\mathrm{D}_\leftrightarrow^\mathit{sys} = 0.07$). Being able to link templates is hence more likely using this reconstruction function than utilizing the scores output by the $\mathit{PIC}$. 

\textbf{Hamming weights function} (Fig.~\ref{fig:scoresAttack:hw}): a function to link templates is proposed in \cite{hermans14bloomDoom}, where templates protected with different keys and belonging to the same instance are shown to have similar Hamming Weights. Therefore, this linkage function will evaluate the Hamming Weight difference between the given protected templates: $s = LS \left( \mathbf{T}_1, \mathbf{T}_2\right) = \vert \mathit{HW}\left(\mathbf{T}_1\right) - \mathit{HW}\left(\mathbf{T}_2\right)\vert$. In this case, only knowledge of the templates is required. 

For this function, the global linkability level is the very similar to the level achieved for the $\mathit{PIC}$ score: $\mathrm{D}_\leftrightarrow^\mathit{sys} = 0.07$. This means that this linkage function cannot extract any further information about whether two templates $\mathbf{T}_1$ and $\mathbf{T}_2$ conceal the same instance. More in detail, only for a small subset of the scores ($s\in [-0.01, 0.01]$) it is slightly more likely that both templates conceal the same instance, and therefore $\mathrm{D}_\leftrightarrow\left( s\right) \in (0, 0.2]$. Since those scores are the most probable (i.e., they yield the highest values in $p\left(s | H_{m}\right) $), the global linkability of the system raises to a value of $\mathrm{D}_\leftrightarrow^\mathit{sys} = 0.08$. 

\textbf{XOR function} (Fig.~\ref{fig:scoresAttack:xor}): in the original concept of Bloom filter-based template protection \cite{RathgebICB}, unlinkability was provided with an XOR operation. However, due to its linearity, templates protected with different keys can be linked exploiting such linearity by $i)$ permuting one of the templates ($\mathbf{T}_2^{\mathrm{perm}}$) and $ii)$ finding the Hamming Distance with respect to the first template: $s = LS\left( \mathbf{T}_1, \mathbf{T}_2\right) = \mathit{HD}\left(\mathbf{T}_1, \mathbf{T}_2^{\mathrm{perm}}\right)$. In this case, only knowledge of the templates is required. 

For this function, the global linkability level is again the very similar to the level achieved for the $\mathit{PIC}$ score. In fact, an even smaller portion of scores on the tails of the distributions allow a successful linkage of the templates with almost all certainty, since $p\left(s | H_{nm}\right) = 0$ and $p\left(s | H_{m}\right) > 0$. Accordingly, $\mathrm{D}_\leftrightarrow\left( s\right) =1$ for those scores. Since those scores yield very low values for $p\left(s | H_{m}\right)$, the chances of obtaining those scores are low, hence showing a global linkability for the system of only $\mathrm{D}_\leftrightarrow^\mathit{sys} = 0.06$.

Finally, following the evaluation guidelines proposed in Sect.~\ref{sec:protocol} and taking into account the four linkage functions analysed (the $\mathit{PIC}$ score in Sect.~\ref{sec:results:PIC} and the three functions evaluated in the present section), we can conclude that the global linkability value of the system is $\mathrm{D}_\leftrightarrow^\mathit{sys} = \max{\{0.07, 0.25, 0.08, 0.06\}} = 0.25$.

\subsection{Analysis of Parameter $\omega$}
\label{sec:exp:omega}

\begin{figure}[t]
\centering
\begin{subfigure}[b]{0.99\linewidth}
\centering
 \centerline{\epsfig{figure=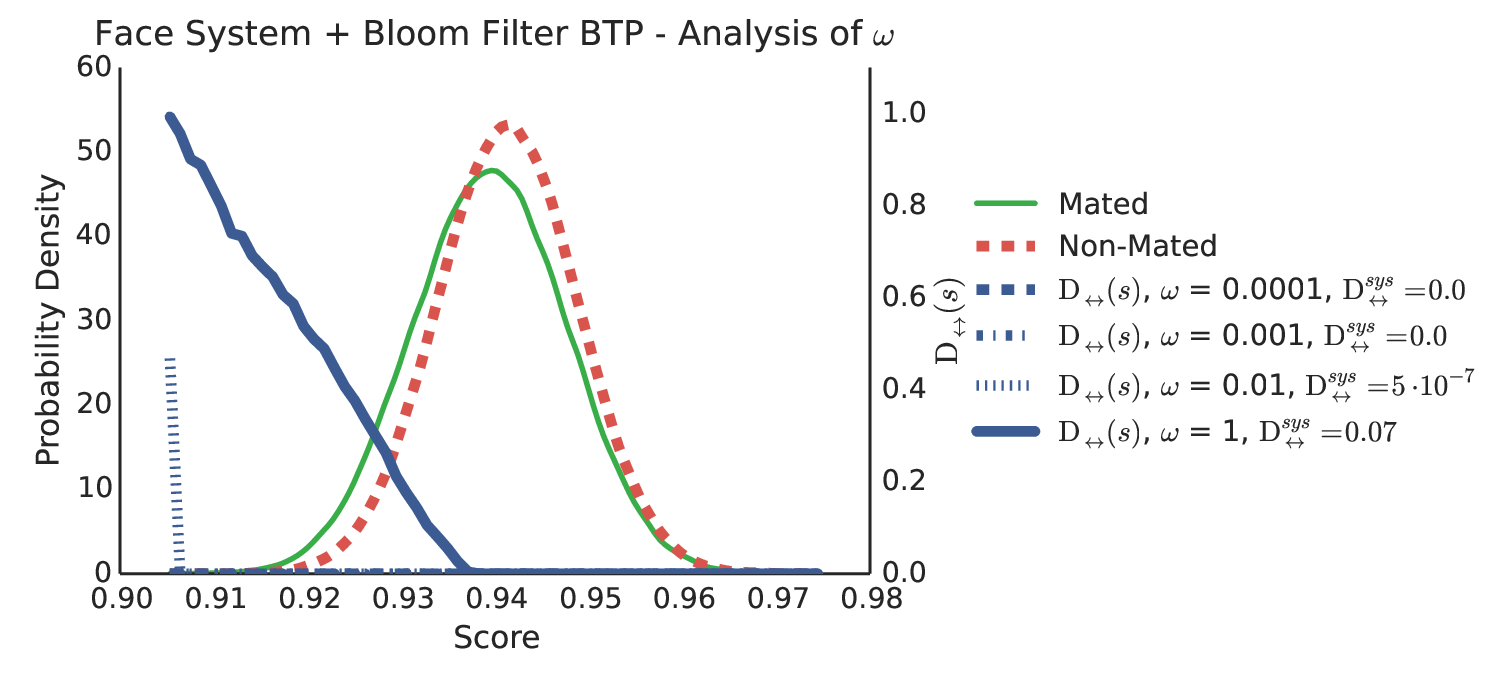,width=.99\linewidth}}\caption{}\label{fig:omega:face}
\end{subfigure}
\begin{subfigure}[b]{0.99\linewidth}
\centering
 \centerline{\epsfig{figure=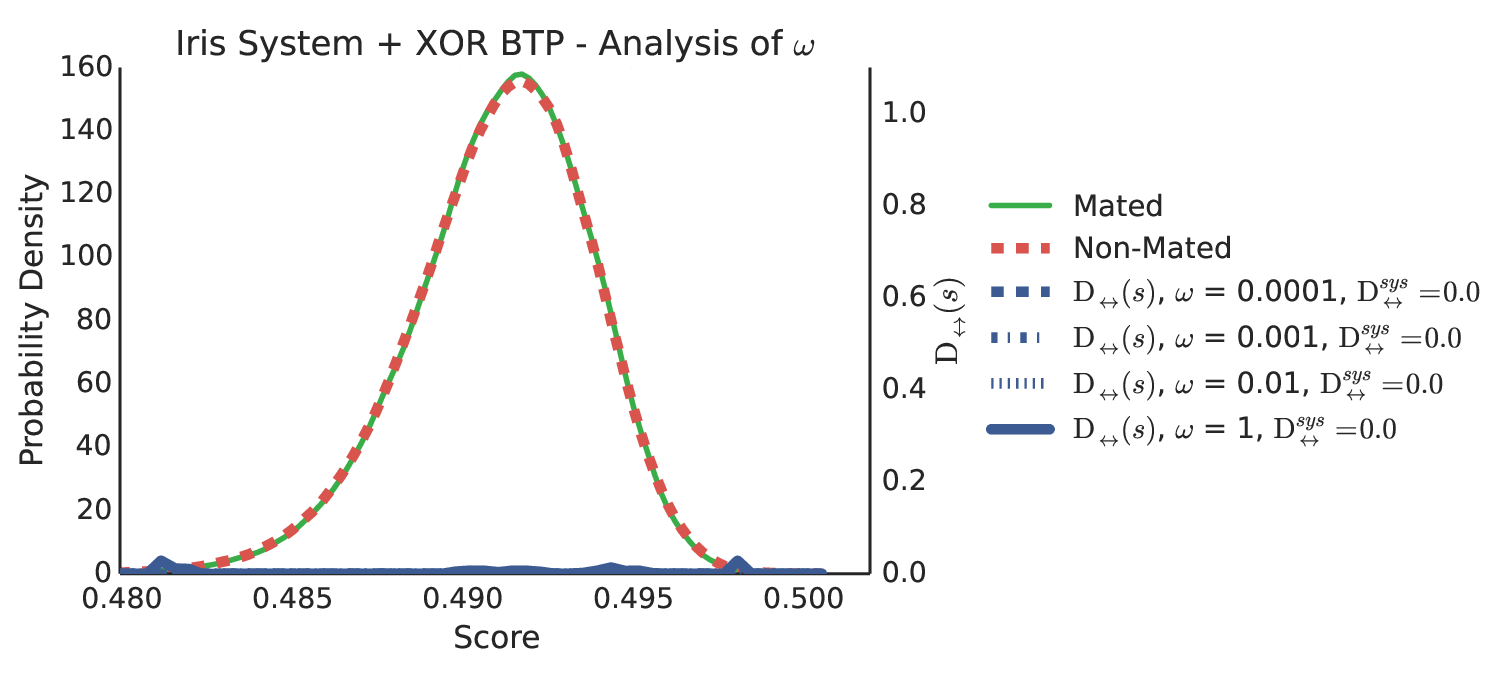,width=.99\linewidth}}\caption{}\label{fig:omega:iris}
\end{subfigure}
\caption{Unlinkability analysis of the facial and iris based BTP systems for different values of $\omega$.} \label{fig:omega}\vspace{-0.4cm}
\end{figure}

In addition to evaluating different linkage functions, the present framework allows the analysis of different scenarios in terms of the a priori probabilities $p\left( H_m\right)$ and $p\left( H_{nm}\right)$. This can be done varying its ratio, $\omega$ (see Sect.~\ref{sec:metric:score} for the definition of this parameter). To that end, four different values $\omega = \left\lbrace 0.0001, 0.001, 0.01, 1\right\rbrace$ have been used in Fig.~\ref{fig:omega} to analyse the \textit{PIC} linkage scores of the facial and iris based BTP schemes. 

As it may be observed, the higher $\omega$ is, the higher the value of $\mathrm{D}_\leftrightarrow\left( s \right)$ for each score $s$ with $p\left( s | H_m \right) > 0$ is. This is due to the fact that the ratio $\left( LR\left( s\right) \cdot \omega \right) / \left( 1 + LR\left( s\right) \cdot \omega \right)$ increases with $\omega$. Regarding the unlinkability property, this also reflects the fact that, for a higher $\omega$, the attacker knows that the probability of having a mated comparison is higher (a higher $\omega$ implies a smaller number of enrolled subjects), and therefore the probabilities of linking the subjects increases.

Finally, such increase in $\mathrm{D}_\leftrightarrow\left( s \right)$ also has an impact on the global linkability of the system: for $\omega = 1$, $\mathrm{D}_\leftrightarrow^\mathit{sys}$ reaches its maximum value.

\begin{figure*}[t]
\centering
\begin{subfigure}[b]{0.29\linewidth}
 \centering 
 \centerline{\epsfig{figure=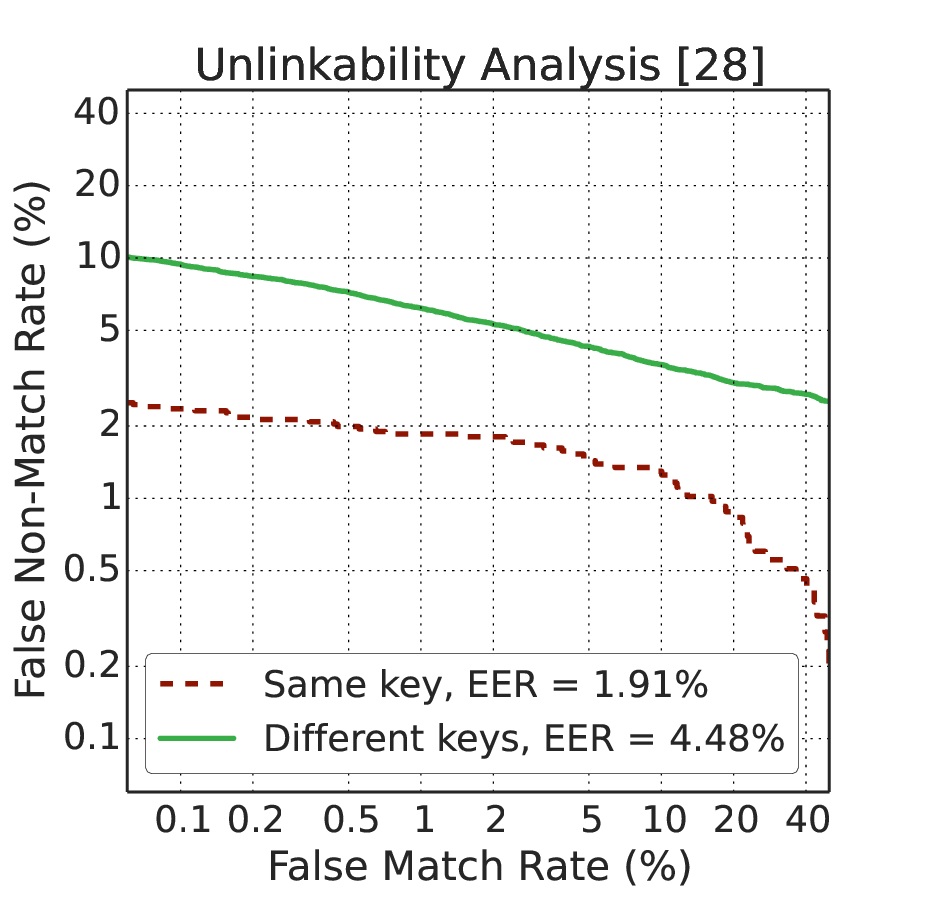,width=.99\linewidth}}\caption{DET curves \cite{kelkboom2011crossmatchFuzzy} \\\hspace{\textwidth}\phantom{ccccccccccccccccccccccccc}}\label{fig:unlinkFV:DET}
\end{subfigure}
\begin{subfigure}[b]{0.29\linewidth}
 \centering 
 \centerline{\epsfig{figure=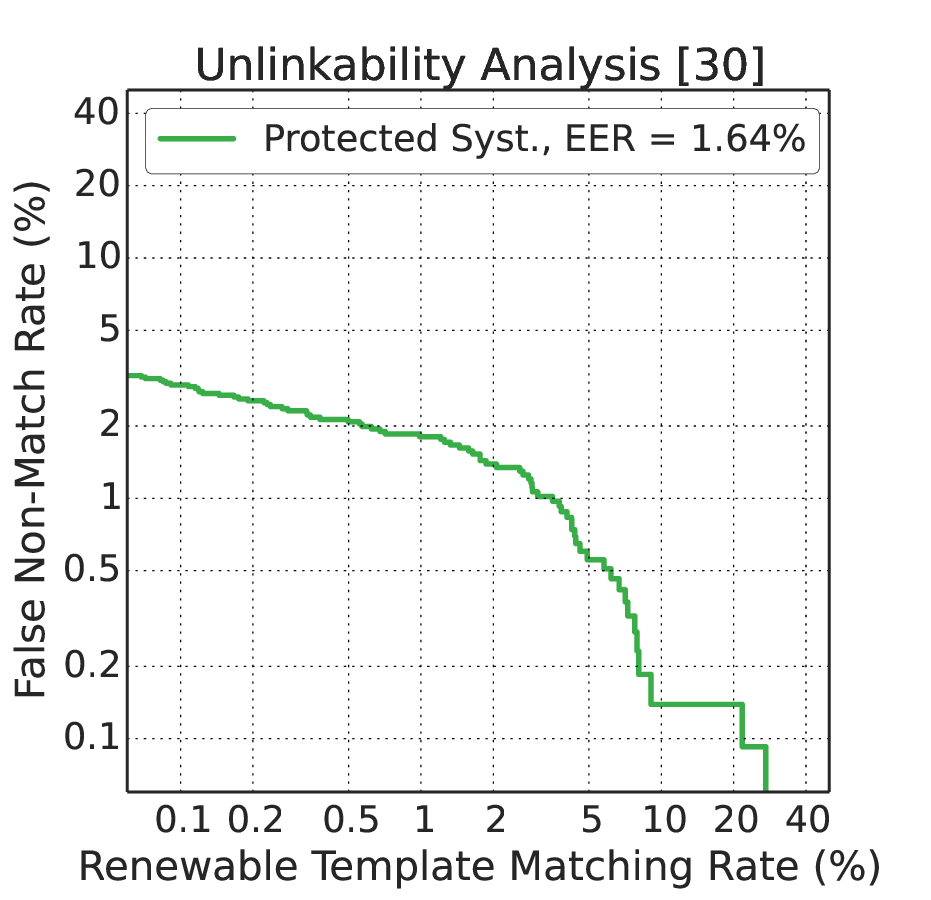,width=.99\linewidth}}\caption{Renewable Template Matching Rate \cite{piciucco2016cancelableFV}}\label{fig:unlinkFV:Ema}
\end{subfigure}
\rulesep
\begin{subfigure}[b]{0.38\linewidth}
 \centering 
 \centerline{\epsfig{figure=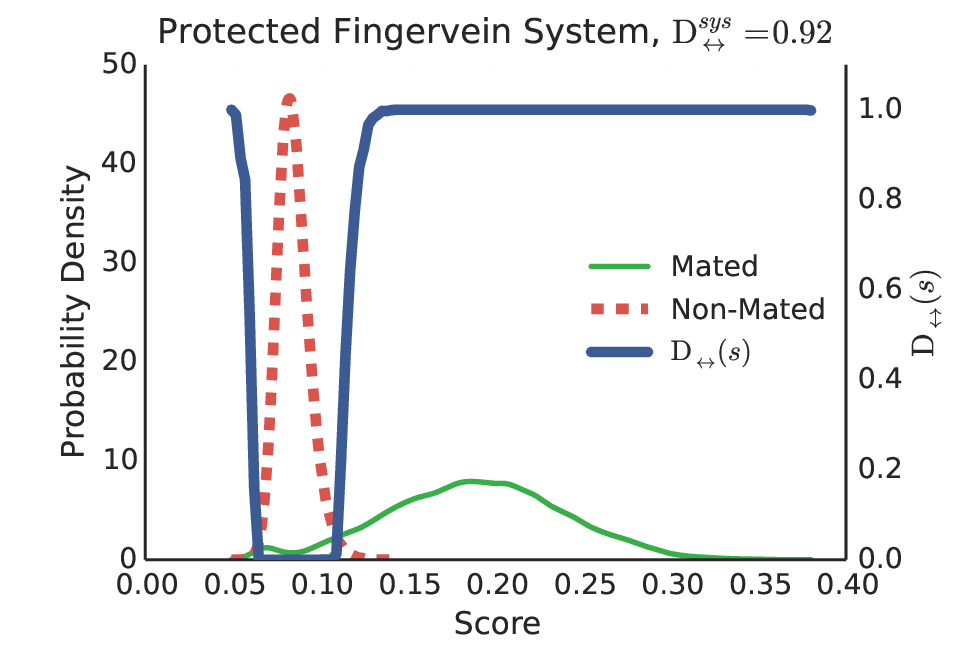,width=.99\linewidth}}\caption{Proposed metrics: protected system \\\hspace{\textwidth}\phantom{cccc}}\label{fig:unlinkFV:oursProt}
\end{subfigure}
\caption{Unlinkability evaluation of the fingervein system using: $i)$ previously proposed metrics: left of the vertical line, figures (a) and (b), and $ii)$ the framework proposed in this paper, right of the vertical line, figure (c).} \label{fig:unlinkFV}\vspace{-0.4cm}
\end{figure*}

\subsection{Advantages over Previously Proposed Metrics}
\label{sec:exp:comp}

In this last set of experiments, we show how the proposed framework is able to determine that a system is linkable even when other existing metrics fail to unveil this vulnerability. To that end, we analyse the fingervein system with two of the general unlinkability metrics that have been proposed in the state-of-the-art \cite{kelkboom2011crossmatchFuzzy,piciucco2016cancelableFV} and with the framework proposed in this article, for the $\mathit{PIC}$ linkage function. It should be highlighted that, although the approaches presented in \cite{kelkboom2011crossmatchFuzzy,piciucco2016cancelableFV} are based on the analysis of the DET curves, in the end unlinkability is considered as a binary value: systems are either linkable or unlinkable. The results are depicted in Fig.~\ref{fig:unlinkFV}. 

In the first approach proposed in \cite{kelkboom2011crossmatchFuzzy} (see Fig.~\ref{fig:unlinkFV:DET}), regular DET curves are depicted in two scenarios: $i)$ the recognition accuracy evaluation (dashed red), where both mated and non-mated scores are computed on templates protected with a single key, and $ii)$ the unlinkability analysis (solid green), where mated scores stem from templates extracted from the \textit{same} instance and protected with \textit{different} keys, and non-mated scores from templates extracted from \textit{different} instances and protected with \textit{different} keys. It should be noted that in \cite{nagar2010btpSecurity}, the FMR of the second scenario is denoted Cross-Match Rate (CMR) and the FNMR is denoted False Cross Match Rate (FCMR). As it may be observed, the error rates increase for the unlinkability analysis with respect to the accuracy analysis. In particular, the EER increases from 1.91\% to 4.48\%. Therefore, the task of discriminating templates extracted from mated and non-mated instances is harder when the templates are protected with different keys. According to the analysis carried out in \cite{buhan2009indistinguishabilityBCS,kelkboom2011crossmatchFuzzy,nagar2010btpSecurity}, we should hence conclude that the protected system is unlinkable.


In the second approach proposed in \cite{piciucco2016cancelableFV}, the DET-like curve comparing the FNMR of templates protected with a single key and the RTMR of templates protected with different keys is depicted in Fig.~\ref{fig:unlinkFV:Ema}. In this case, the curve shows a similar behaviour to the accuracy curve shown in Fig.~\ref{fig:unlinkFV:DET} in dashed red. Therefore, as the authors suggest in \cite{piciucco2016cancelableFV}, the task of discriminating templates protected with different keys is as hard as achieving a false match with a random template, thereby making the protected system unlinkable.



Finally, the protected system using different keys is analysed in Fig.~\ref{fig:unlinkFV:oursProt} with the metrics proposed in the present article. As it may be observed, $\mathrm{D}_\leftrightarrow^\mathit{sys}$ yields a value of 0.92, very close to 1, thereby indicating that the templates are almost fully linkable. This is corroborated by the mated and non-mated distributions, which are almost fully separable. As a consequence, $\mathrm{D}_\leftrightarrow\left( s \right) = 1$ for almost the entire domain of scores for both systems. That means that the simple linkage function that uses the similarity scores of the $\mathit{PIC}$ is enough to link protected templates, hence contradicting the results of the evaluations carried out following \cite{kelkboom2011crossmatchFuzzy,piciucco2016cancelableFV}.


\section{Conclusions}
\label{sec:conc}

We have proposed in the present article two new quantitative measures for the unlinkability analysis of biometric templates, which can be applied to any biometric template protection scheme, regardless of its overall strategy (i.e., cancelable biometrics, cryptobiometrics or encrypted system). On the one hand, $D_\leftrightarrow\left(s \right)$ provides a score-wise analysis of the linkability of the templates, in order to carry out a thorough and detailed evaluation of the templates' unlinkability. On the other hand, $D_\leftrightarrow^\mathit{sys}$ evaluates the system as a whole, thereby allowing a benchmark of the linkability of different systems. Furthermore, the necessary steps towards a complete unlinkability evaluation have been proposed in order to develop a full security benchmark for biometric template protection schemes, key for the further deployment of biometric systems, as stated in \cite{rane14standardBPT}. We therefore believe that the proposed framework will contribute to the advancement of biometric technologies in the future. To that end, and with the aim to make the article reproducible, an implementation of the metrics has been made public through the da/sec website and the da/sec Github account. 

As with any other security oriented property, linkability is not an intrinsic property of the system. Instead, it is directly related to particular threats in the form of $i)$ linkage functions and $ii)$ security models assumed, which must be analysed on a one by one basis. This is therefore the approach followed in the proposed framework. Both metrics evaluate in a quantitative manner the degree of linkability of protected biometric templates with respect to a given linkage function, which may exploit any particular weakness of the system at hand. 

For the evaluation, only access to the scores obtained from the corresponding linkage functions is needed, without assumptions on the input data or the nature of the templates and systems. We deem that such requirement is straightforward for both system developers and evaluators, as no access to inner modules or communication channels of the system is required. 

A direct consequence of this sole requirement of the framework is the general applicability of the metrics. Contrary to some previously proposed unlinkability metrics, where assumptions on the underlying algorithms of the system or the data were made, there is no restriction on the type of systems which can be analysed, and the number of linkage functions for each system. The only requirement is the use of Lebesgue integrable linkage functions. We believe this is a minor limitation, since the majority of usual linkage functions meet that condition, and other non-continuous functions can be mapped to distance-based functions for the analysis. This has been shown by evaluating four previously proposed biometric template protection systems based on different characteristics and protection strategies, as well as several linkage functions.

In addition, and with respect to the previously proposed approaches for unlinkability assessment, it has been shown that the proposed metrics can reveal linkability vulnerabilities concealed to other metrics. Furthermore, our metrics do so by providing not only local information ($\mathrm{D}_\leftrightarrow\left( s \right)$) but also a global measure ($\mathrm{D}_\leftrightarrow^\mathit{sys}$), both taking into account the continuous nature of linkability (i.e., instead of a binary decision, different degrees of unlinkability can be achieved). As a consequence, the framework provides a fair benchmark of several systems and linkage functions as long as the same evaluation protocol is followed (e.g., same data are used).

Finally, it should be noted that the proposed framework can be applied not only to the evaluation of biometric recognition schemes, but also to other fields where privacy protection is key. For instance, whereas user profiling in social media can bring benefits such as content \cite{hannon2010profilingRecommenderSyst} or travel recommendation \cite{memon2015travelRecommendProfiling}, sensitive information can also be recovered out of seemingly anonymous data \cite{mitrou2014socialMediaProfiling}. As stated in \cite{mitrou2014socialMediaProfiling}, a direct consequence of this fact is that \lq\lq individuals may be confronted with social exclusion, prejudice and discrimination risks both in their workplace and in their social environment''. The proposed framework can be therefore used to analyse how vulnerable the information stored in the cloud is to such profiling (i.e., linking) activities, and further help to prevent these undesired practices.

\appendices

\section{Mathematical Proofs}
\label{sec:annex}

We present here the proofs of the properties of both linkability measures.

\subsection{Local Measure $\mathrm{D}_{\leftrightarrow} \left( s \right)$ Properties}
\label{sec:annex:score}

\textbf{Domain}. Since $ LR\left(s\right) $ is defined over the whole domain of $s$ scores, $\mathrm{D}_{\leftrightarrow}  \left( s \right)$ is defined for any two $p\left( s | H_{m} \right)$ and $p\left( s | H_{nm} \right)$ distributions.

\textbf{Continuity}. Additionally, $\mathrm{D}_{\leftrightarrow} \left( s \right)$ is continuous since it is piecewise continuous and it is also continuous at $LR \left( s \right) \cdot \omega = 1$:
\begin{align}
\lim_{LR \left( s \right) \cdot \omega \to 1^-} &\mathrm{D}_{\leftrightarrow} \left( s \right) = \lim_{LR \left( s \right) \cdot \omega \to 1^-} 0 = 0
\\
\begin{split}
\lim_{LR \left( s \right) \cdot \omega \to 1^+} &\mathrm{D}_{\leftrightarrow} \left( s \right) = \lim_{LR \left( s \right) \cdot \omega \to 1^+} 2\frac{LR \left( s \right) \cdot \omega}{1 + LR\left( s \right) \cdot \omega} - 1= 0
\end{split}
\end{align}

\textbf{Range}. Furthermore, $\mathrm{D}_{\leftrightarrow} \left( s \right)$ is bounded in $[0,1]$:
\begin{align}
\lim_{LR\left(s\right) \cdot \omega \to 0} &\mathrm{D}_{\leftrightarrow} \left( s \right) = \lim_{LR \left( s \right) \cdot \omega \to 0} 0 = 0
\\
\begin{split}
\lim_{LR\left(s\right) \cdot \omega \to +\infty} &\mathrm{D}_{\leftrightarrow} \left( s \right) =
\\
& \lim_{LR\left(s\right) \cdot \omega \to +\infty} 2\frac{LR \left( s \right) \cdot \omega}{1 + LR\left( s \right) \cdot \omega} - 1 \underset{\mathrm{L'H\hat{o}pital}}{\to}
\\
& \lim_{LR\left(s\right) \cdot \omega \to +\infty} 2\frac{1}{1} - 1 = 1
\end{split}
\end{align}

\textbf{Monotonicity}. $\mathrm{D}_{\leftrightarrow} \left( s \right)$ is a monotonically increasing function, since its derivative is always non-negative:
\begin{equation}\label{eq:unlinkDer}
\mathrm{D}_{\leftrightarrow}'  \left( s \right) = 
\begin{cases}
0 &\text{if } LR\left(s\right) \cdot \omega \le 1
\\
2 \frac{\omega}{\left( 1 +  LR\left(s\right) \cdot \omega \right)^2} > 0 &\text{if } LR\left(s\right)\cdot \omega > 1
\end{cases}
\end{equation}
being the second term positive since all figures involved are positive numbers.
As a consequence of its monotonicity, $\mathrm{D}_{\leftrightarrow} \left( s \right)$ provides appropriate increasing values for the evaluation of a monotonic property such as templates linkability.

\subsection{Global Measure $\mathrm{D}_{\leftrightarrow}^{\mathit{sys}}$ Properties}
\label{sec:annex:global}

\textbf{Properly defined}. Since $\mathrm{D}_{\leftrightarrow} \left( s \right)$ and $p\left( s \vert H_{m} \right)$ are continuous functions, their product is also continuous. As a consequence, by the Riemann-Lebesgue theorem \cite{apostol74analysis}, the product is integrable, hence being $\mathrm{D}_{\leftrightarrow}^{\mathit{sys}}$ properly defined. This property also holds for discrete functions, since they are also Lebesgue integrable.

\textbf{Range}. Let us now prove that $\mathrm{D}_{\leftrightarrow}^{\mathit{sys}} \in [0, 1]$. On the one hand, since $p\left( s \vert H_{m} \right)$ is a probability density, which, integrated over $[s_{min}, s_{max}]$ yields an area of one, we have
\begin{equation}\label{eq:unlinkSystemLimit1}
\mathrm{D}_{\leftrightarrow}^{\mathit{sys}} = \int  \mathrm{D}_{\leftrightarrow}  \left( s \right)  \cdot p\left( s \vert H_{m} \right)  \mathrm{d}s \le \int 1 \cdot p\left( s \vert H_{m} \right)  \mathrm{d}s = 1
\end{equation}

Similarly,
\begin{equation}\label{eq:unlinkSystemLimit0}
\mathrm{D}_{\leftrightarrow}^{\mathit{sys}} = \int \mathrm{D}_{\leftrightarrow}  \left( s \right)  \cdot p\left( s \vert H_{m} \right)  \mathrm{d}s \ge \int 0 \cdot p\left( s \vert H_{m} \right)  \mathrm{d}s = 0
\end{equation}

\section*{Acknowledgements}

This work was supported by the German Federal Ministry of Education and Research (BMBF) as well as by the Hessen State Ministry for Higher Education, Research and the Arts (HMWK) within the Center for Research in Security and Privacy (CRISP, \url{www.crisp-da.de}).

\bibliographystyle{IEEEtran}
\bibliography{bib/references}

\begin{thebibliography}{10}
\providecommand{\url}[1]{#1}
\csname url@samestyle\endcsname
\providecommand{\newblock}{\relax}
\providecommand{\bibinfo}[2]{#2}
\providecommand{\BIBentrySTDinterwordspacing}{\spaceskip=0pt\relax}
\providecommand{\BIBentryALTinterwordstretchfactor}{4}
\providecommand{\BIBentryALTinterwordspacing}{\spaceskip=\fontdimen2\font plus
\BIBentryALTinterwordstretchfactor\fontdimen3\font minus
  \fontdimen4\font\relax}
\providecommand{\BIBforeignlanguage}[2]{{%
\expandafter\ifx\csname l@#1\endcsname\relax
\typeout{** WARNING: IEEEtran.bst: No hyphenation pattern has been}%
\typeout{** loaded for the language `#1'. Using the pattern for}%
\typeout{** the default language instead.}%
\else
\language=\csname l@#1\endcsname
\fi
#2}}
\providecommand{\BIBdecl}{\relax}
\BIBdecl

\bibitem{jain07nature}
A.~K. Jain, ``Technology: Biometric recognition,'' \emph{Nature}, vol. 449, pp.
  38--49, 2007.

\bibitem{indianUID}
\BIBentryALTinterwordspacing
{Government of India}, ``Unique identification authority of india,'' 2012.
  [Online]. Available: \url{https://uidai.gov.in/}
\BIBentrySTDinterwordspacing

\bibitem{SmartBorders}
\BIBentryALTinterwordspacing
{European Comission}, ``Smart borders,'' 2013. [Online]. Available:
  \url{http://ec.europa.eu/dgs/home-affairs/what-we-do/policies/borders-and-visas/smart-borders/index_en.htm}
\BIBentrySTDinterwordspacing

\bibitem{euregulation16}
{European Council}, ``{Regulation of the European Parliament and of the Council
  on the protection of individuals with regard to the processing of personal
  data and on the free movement of such data (General Data Protection
  Regulation)},'' 04 2016.

\bibitem{campisi13secPrivacyBio}
P.~Campisi, Ed., \emph{Security and Privacy in Biometrics}.\hskip 1em plus
  0.5em minus 0.4em\relax Springer, 2013.

\bibitem{patel15CancelableBioSurvey}
V.~M. Patel, N.~Ratha, and R.~Chellappa, ``Cancelable biometrics: A review,''
  \emph{IEEE Signal Proc. Magazine}, vol.~32, no.~5, pp. 54--65, 2015.

\bibitem{ISO-IEC-24745:2011}
{ISO/IEC JTC1 SC27 Security Techniques}, \emph{ISO/IEC 24745:2011. Information
  Technology - Security Techniques - Biometric Information Protection}, ISO,
  2011.

\bibitem{ISO-IEC-30136-2017}
{ISO/IEC JTC1 SC37 Biometrics}, \emph{{ISO/IEC FDIS} 30136, Performance Testing
  of Biometric Template Protection Schemes}, International Organization for
  Standardization, 2017.

\bibitem{rane14standardBPT}
S.~Rane, ``Standardization of biometric template protection,'' \emph{IEEE
  Multimedia}, vol.~21, no.~4, pp. 94--99, 2014.

\bibitem{BUludag04a}
U.~Uludag, S.~Pankanti \emph{et~al.}, ``Biometric cryptosystems: issues and
  challenges,'' \emph{Proc. of the IEEE}, vol.~92, no.~6, pp. 948--960, 2004.

\bibitem{BRatha01a}
N.~Ratha, J.~Connell, and R.~Bolle, ``Enhancing security and privacy in
  biometrics-based authentication systems,'' \emph{IBM Systems Journal},
  vol.~40, no.~3, pp. 614--634, 2001.

\bibitem{Rathgeb11e}
C.~Rathgeb and A.~Uhl, ``A survey on biometric cryptosystems and cancelable
  biometrics,'' \emph{EURASIP Journal on Information Security}, vol. 2011,
  no.~3, 2011.

\bibitem{Ferrara14a}
M.~Ferrara, D.~Maltoni, and R.~Cappelli, ``A two-factor protection scheme for
  {MCC} fingerprint templates,'' in \emph{Proc. BIOSIG}, 2014.

\bibitem{champod2000LRspeaker}
C.~Champod and D.~Meuwly, ``The inference of identity in forensic speaker
  recognition,'' \emph{Speech Comm.}, vol.~31, no.~2, pp. 193--203, 2000.

\bibitem{bazen2004LRverification}
A.~Bazen and R.~Veldhuis, ``Likelihood-ratio-based biometric verification,''
  \emph{IEEE Trans. on Circuits and Systems for Video Technology}, vol.~14,
  no.~1, pp. 86--94, 2004.

\bibitem{joaquin06LRspeaker}
J.~Gonzalez-Rodriguez, A.~Drygajlo \emph{et~al.}, ``Robust estimation,
  interpretation and assessment of likelihood ratios in forensic speaker
  recognition,'' \emph{Comp. Speech \&\ Language}, vol.~20, no.~2, pp.
  331--355, 2006.

\bibitem{ali2013LRface}
T.~Ali, L.~Spreeuwers, R.~Veldhuis, and D.~Meuwly, ``Effect of calibration data
  on forensic likelihood ratio from a face recognition system,'' in \emph{Proc.
  Int. Conf. on Biometrics: Theory, Applications, and Systems, BTAS}, 2013, pp.
  1--8.

\bibitem{Ramos2017LRfp}
D.~Ramos, R.~Haraksim, and D.~Meuwly, ``Likelihood ratio data to report the
  validation of a forensic fingerprint evaluation method,'' \emph{Data in
  Brief}, vol.~10, pp. 75--92, 2017.

\bibitem{linnartz2003shieldingFunct}
J.-P. Linnartz and P.~Tuyls, ``New shielding functions to enhance privacy and
  prevent misuse of biometric templates,'' in \emph{Proc. Int. Conf. on
  Audio‑ and Video‑Based Biometric Person Authentication, AVBPA}, 2003, pp.
  393--402.

\bibitem{dodis2004bioKeys}
Y.~Dodis, L.~Reyzin, and A.~Smith, ``How to generate strong keys from
  biometrics and other noisy data,'' in \emph{Proc. Eurocrypt}, 2004, pp.
  523--540.

\bibitem{buhan2007fuzzyContinuous}
I.~Buhan, J.~Doumen, P.~Hartel, and R.~Veldhuis, ``Fuzzy extractors for
  continuous distributions,'' in \emph{Proc. ACM Int. Conf. on Computational
  Science, ICCS}, 2007, pp. 353--355.

\bibitem{simoens2012framework}
K.~Simoens, J.~Bringer, H.~Chabanne, and S.~Seys, ``A framework for analyzing
  template security and privacy in biometric authentication systems,''
  \emph{IEEE Trans. on Information Forensics and Security}, vol.~7, no.~2, pp.
  833--841, 2012.

\bibitem{bringer15secAnalysisBF}
J.~Bringer, C.~Morel, and C.~Rathgeb, ``Security analysis of bloom filter-based
  iris biometric template protection,'' in \emph{Proc. Int. Conf. on
  Biometrics, ICB}, 2015, pp. 527--534.

\bibitem{kholmatov2008corrAttackFV}
A.~Kholmatov and B.~Yanikoglu, ``Realization of correlation attack against the
  fuzzy vault scheme,'' in \emph{Proc. Electronic Imaging}, 2008, pp.
  68\,190O--68\,190O.

\bibitem{simoens2009indistinguishabilityBioSketches}
K.~Simoens, P.~Tuyls, and B.~Preneel, ``Privacy weaknesses in biometric
  sketches,'' in \emph{Proc. IEEE Signal Processing}, 2009, pp. 188--203.

\bibitem{buhan2009indistinguishabilityBCS}
I.~Buhan, J.~Breebaart, J.~Guajardo \emph{et~al.}, ``A quantitative analysis of
  indistinguishability for a continuous domain biometric cryptosystem,'' in
  \emph{Proc. Int. Conf. on Data Privacy Management and Autonomous Spontaneous
  Security, DPM/SETOP}, 2009, pp. 78--92.

\bibitem{buhan2010indistinguishabilityFuzzy}
I.~Buhan, J.~Merchan, and E.~Kelkboom, ``Efficient strategies for playing the
  indistinguishability game for fuzzy sketches,'' in \emph{Proc. Int. Workshop
  on Information Forensics and Security, WIFS}, 2010.

\bibitem{kelkboom2011crossmatchFuzzy}
E.~J. Kelkboom, J.~Breebaart, T.~A. Kevenaar, I.~Buhan, and R.~N. Veldhuis,
  ``Preventing the decodability attack based cross-matching in a fuzzy
  commitment scheme,'' \emph{IEEE Trans. on Information Forensics and
  Security}, vol.~6, no.~1, pp. 107--121, 2011.

\bibitem{nagar2010btpSecurity}
A.~Nagar, K.~Nandakumar, and A.~K. Jain, ``Biometric template transformation: a
  security analysis,'' in \emph{Proc. SPIE 7541}.\hskip 1em plus 0.5em minus
  0.4em\relax International Society for Optics and Photonics, 2010, p. 75410O.

\bibitem{piciucco2016cancelableFV}
E.~Piciucco, E.~Maiorana \emph{et~al.}, ``Cancelable biometrics for finger vein
  recognition,'' in \emph{Proc. SPLINE}.\hskip 1em plus 0.5em minus 0.4em\relax
  IEEE, 2016, pp. 1--5.

\bibitem{rua12BTPandUBMsign}
E.~A. Rua, E.~Maiorana, J.~L.~A. Castro, and P.~Campisi, ``Biometric template
  protection using universal background models: An application to online
  signature,'' \emph{IEEE Trans. on Information Forensics and Security},
  vol.~7, no.~1, pp. 269--282, 2012.

\bibitem{wang2014cancelableFpConv}
S.~Wang and J.~Hu, ``Design of alignment-free cancelable fingerprint templates
  via curtailed circular convolution,'' \emph{Pattern Recognition}, vol.~47,
  no.~3, pp. 1321--1329, 2014.

\bibitem{kullback51KLdivergence}
S.~Kullback and R.~A. Leibler, ``On information and sufficiency,'' \emph{The
  Annals of Mathematical Statistics}, vol.~22, no.~1, pp. 79--86, 1951.

\bibitem{ISO-IEC-2382-37:2012}
{ISO/IEC TC JTC1 SC37 Biometrics}, \emph{ISO/IEC 2382-37:2012 Information
  technology -- Vocabulary -- Part 37: Biometrics}, ISO and IEC, 2012.

\bibitem{ferrara12MCCnonInv}
M.~Ferrara, D.~Maltoni, and R.~Cappelli, ``Non-invertible minutia cylinder-code
  representation,'' \emph{IEEE Trans. on Information Forensics and Security},
  vol.~7, no.~6, pp. 1727--1737, 2012.

\bibitem{mansfield02best}
A.~Mansfield and J.~Wayman, ``Best practices in testing and reporting
  performance of biometric devices,'' CESG Biometrics Working Group, Tech.
  Rep., August 2002, (http://www.cesg.gov.uk/).

\bibitem{Uhl12a}
A.~Uhl and P.~Wild, ``Weighted adaptive hough and ellipsopolar transforms for
  real-time iris segmentation,'' in \emph{Proc. Int. Cong. on Biometrics, ICB},
  2012, pp. 1--8.

\bibitem{masek03irisSystem}
L.~Masek and P.~Kovesi, ``{MATLAB} source code for a biometric identification
  system based on iris patterns,'' Master's thesis, University of Western
  Australia, 2003.

\bibitem{zuo08verticalShiftIris}
J.~Zuo, N.~K. Ratha, and J.~H. Connell, ``Cancelable iris biometric,'' in
  \emph{Proc. ICPR}.\hskip 1em plus 0.5em minus 0.4em\relax IEEE, 2008, pp.
  1--4.

\bibitem{marcos13dtw}
M.~Martinez-Diaz, J.~Fierrez \emph{et~al.}, ``Mobile signature verification:
  Feature robustness and performance comparison,'' \emph{IET Biometrics},
  vol.~3, no.~4, pp. 267--277, 2014.

\bibitem{galbally08GlobalFeatSelection}
J.~Galbally, J.~Fierrez, and J.~Ortega-Garcia, ``Performance and robustness: a
  trade-off in dynamic signature verification,'' in \emph{Proc. Int. Conf. on
  Acoustics, Speech, and Signal Processing, ICASSP}, 2008.

\bibitem{marta16HEGLobalFeats}
M.~Gomez-Barrero, J.~Galbally, E.~Maiorana, P.~Campisi, and J.~Fierrez,
  ``Implementation of fixed-length template protection based on {Homomorphic
  Encryption} with application to signature biometrics,'' in \emph{Proc. Int.
  Conf. on Computer Vision and Pattern Recognition Workshops, CVPRW}, 2016, pp.
  191--198.

\bibitem{zhang05LGBPHS}
W.~Zhang, S.~Shan \emph{et~al.}, ``Local gabor binary pattern histogram
  sequence ({LGBPHS}): a novel non-statistical model for face representation
  and recognition,'' in \emph{Proc. Int. Conf. on Computer Vision, ICCV},
  vol.~1, 2005, pp. 786--791.

\bibitem{gunther12facereclib}
M.~G{\"u}nther, R.~Wallace, and S.~Marcel, ``An open source framework for
  standardized comparisons of face recognition algorithms,'' in \emph{Proc.
  European Conf. on Computer Vision, ECCV}, ser. LNCS, vol. 7585, 2012, pp.
  547--556.

\bibitem{marta16BFsUnlink}
M.~Gomez-Barrero, C.~Rathgeb, J.~Galbally, C.~Busch, and J.~Fierrez,
  ``Unlinkable and irreversible biometric template protection based on {Bloom}
  filters,'' \emph{Information Sciences}, vol. 370-371, pp. 18--32, 2016.

\bibitem{miura07MCPfingervein}
N.~Miura, A.~Nagasaka, and T.~Muyatake, ``Extraction of finger-vein patterns
  using maximum curvature points in image profiles,'' \emph{IEICE Trans. on
  Information and Systems}, vol.~90, no.~8, pp. 1185--1194, 2007.

\bibitem{ratha01securityPrivacy}
N.~K. Ratha, J.~H. Connell, and R.~M. Bolle, ``Enhancing security and privacy
  in biometrics-based authentication sytems,'' \emph{IBM Systems Journal},
  vol.~40, pp. 614--634, 2001.

\bibitem{biosecure09PAMI}
J.~Ortega-Garcia, J.~Fierrez \emph{et~al.}, ``The multi-scenario
  multi-environment {B}io{S}ecure multimodal database ({BMDB}),'' \emph{IEEE
  Trans. on Pattern Analysis and Machine Intelligence}, vol.~32, pp.
  1097--1111, 2010.

\bibitem{UTFVPDB}
B.~Ton and R.~Veldhuis, ``A high quality finger vascular pattern dataset
  collected using a custom designed capturing device,'' in \emph{Proc. Int.
  Conf. on Biometrics, ICB}, 2013.

\bibitem{simoens12BTPmetrics}
K.~Simoens, B.~Yang \emph{et~al.}, ``Criteria towards metrics for benchmarking
  template protection algorithms,'' in \emph{Proc. ICB}, 2012, pp. 498--505.

\bibitem{hermans14bloomDoom}
J.~Hermans, B.~Mennink, and R.~Peeters, ``When a bloom filter is a doom filter:
  Security assessment of a novel iris biometric,'' in \emph{Proc. BIOSIG},
  2014.

\bibitem{RathgebICB}
C.~Rathgeb, F.~Breitinger, and C.~Busch, ``Alignment-free cancelable iris
  biometric templates based on adaptive bloom filters,'' in \emph{Proc. Int.
  Conf. on Biometrics, ICB}, 2013, pp. 1--8.

\bibitem{hannon2010profilingRecommenderSyst}
J.~Hannon, M.~Bennett, and B.~Smyth, ``Recommending twitter users to follow
  using content and collaborative filtering approaches,'' in \emph{Proc. ACM
  Int. Conf. on Recommender Systems}, 2010, pp. 199--206.

\bibitem{memon2015travelRecommendProfiling}
I.~Memon, L.~Chen, A.~Majid, M.~Lv, I.~Hussain, and G.~Chen, ``Travel
  recommendation using geo-tagged photos in social media for tourist,''
  \emph{Wireless Personal Communications}, vol.~80, no.~4, pp. 1347--1362,
  2015.

\bibitem{mitrou2014socialMediaProfiling}
L.~Mitrou, M.~Kandias, V.~Stavrou, and D.~Gritzalis, ``Social media profiling:
  A panopticon or omniopticon tool?'' in \emph{Proc. Int. Conf. of the
  Surveillance Studies Network, CSSN}, 2014.

\bibitem{apostol74analysis}
T.~Apostol, \emph{Mathematical Analysis}, 1974, pp. 169--172.

\end{thebibliography}

\begin{IEEEbiography}[{\includegraphics[width=1in,height=1.25in,clip,keepaspectratio]{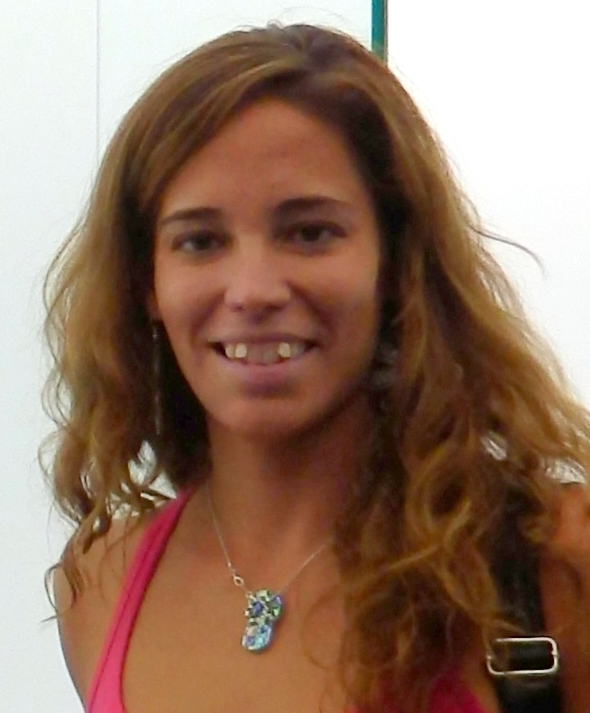}}]%
{Marta Gomez-Barrero}
received her MSc degrees in Computer Science and Mathematics, and her PhD degree in Electrical Engineering, from Universidad Autonoma de Madrid, in 2011 and 2016, respectively. Since 2016 she is a PostDoctoral researcher at the Center for Research in Security and Privacy (CRISP), Germany. Her current research focuses on the development of privacy-enhancing biometric technologies as well as Presentation Attack Detection methods, within the wider fields of pattern recognition and machine learning. She is the recipient of a number of distinctions, including: EAB European Biometric Industry Award 2015, Siew-Sngiem Best Paper Award at ICB 2015, Archimedes Award for young researches from Spanish Ministry of Education in 2013 and Best Poster Award at ICB 2013.
\end{IEEEbiography}

\begin{IEEEbiography}[{\includegraphics[width=1in,height=1.25in,clip,keepaspectratio]{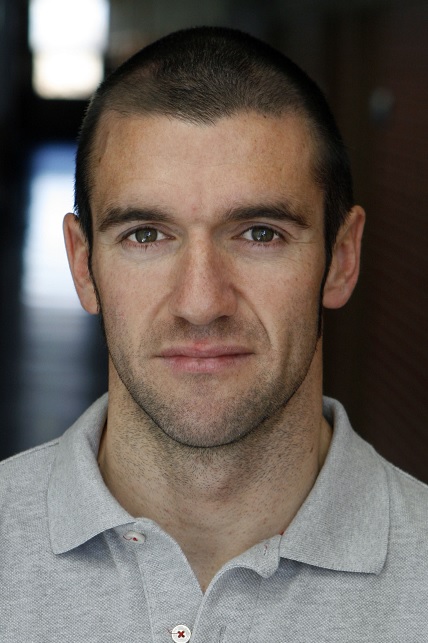}}]%
{Javier Galbally}
received the M.Sc. degree in Electrical Engineering from the Universidad de Cantabria, Spain, in 2005, and the Ph.D. degree in Electrical Engineering from the Universidad Autónoma de Madrid (UAM), Spain, in 2009, where he was an Assistant Professor until 2012. In 2013, he joined the European Commission in DG Joint Research Centre, where he is currently a Post-Doctoral Researcher. His research interests are mainly focused on pattern and biometric recognition, including biometric systems security and vulnerabilities, biometric template protection and inverse biometrics. He is the recipient of a number of distinctions, including the IBM Best Student Paper Award at ICPR 2008, finalist of the EBF European Biometric Research Award 2009, Best Ph.D. Thesis Award 2010 by the UAM, Best Poster Award at IJCB 2013 and 2017, and Best Paper Award at ICB 2015.
\end{IEEEbiography}

\begin{IEEEbiography}[{\includegraphics[width=1in,height=1.25in,clip,keepaspectratio]{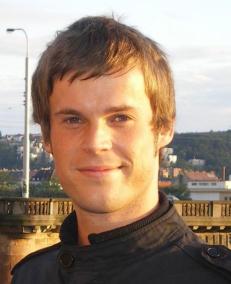}}]%
{Christian Rathgeb}
is a senior researcher with the faculty of computer science, Hochschule Darmstadt, Germany. He is a Principal Investigator in the Center for Research in Security and Privacy (CRISP). His research includes pattern recognition, iris biometrics, and privacy enhancing technologies for biometric systems. He served for various program committees and conferences, journals and magazines as reviewer. He is a member of the European Association for Biometrics (EAB) and a Program Chair of the International Conference of the Biometrics Special Interest Group (BIOSIG). 
\end{IEEEbiography}

\begin{IEEEbiography}[{\includegraphics[width=1in,height=1.25in,clip,keepaspectratio]{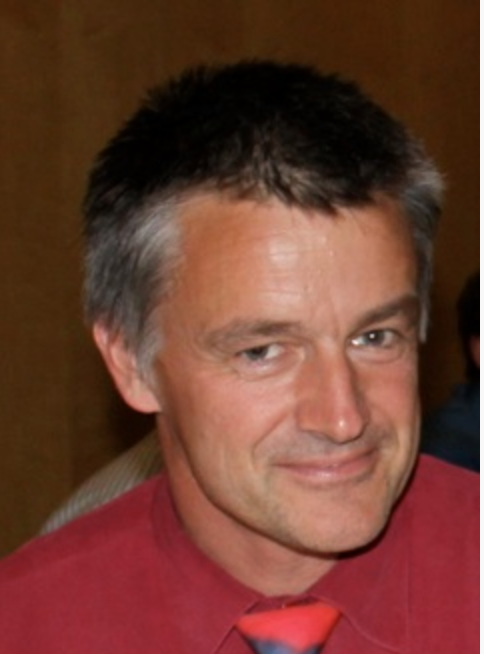}}]%
{Christoph Busch}
received the Diploma degree from the Technical University of Darmstadt (TUD), Darmstadt, Germany, and the Ph.D. degree in computer graphics from TUD, in 1997. He joined the Fraunhofer Institute for Computer Graphics, Darmstadt, in 1997. He is a member of the Faculty of Computer Science and Media Technology with the Norwegian University of Science and Technology, Norway, and holds a joint appointment with the Faculty of Computer Science, Hochschule Darmstadt. Furthermore, he lectures a course on biometric systems with DTU in Copenhagen since 2007. His research includes pattern recognition, multimodal and mobile biometrics, and privacy enhancing technologies for biometric systems. He is Cofounder of the European Association for Biometrics and convener of WG3 in ISO/IEC JTC1 SC37 on Biometrics. He coauthored over 400 technical papers, and has been a speaker at international conferences.
\end{IEEEbiography}

\end{document}